\documentclass[journal]{IEEEtran}
\IEEEoverridecommandlockouts
\usepackage{cite}
\usepackage{amsmath,amssymb,amsfonts}
\usepackage{algorithm}
\usepackage{algorithmicx}
\usepackage{algpseudocode}
\usepackage{graphicx}
\usepackage{tabularx}
\usepackage{multirow}
\usepackage{booktabs}
\usepackage{textcomp}
\usepackage{xcolor}

\newcommand{\R}{\mathbb{R}}

\newcommand{\ba}{\boldsymbol{a}}
\newcommand{\bu}{\boldsymbol{u}}
\newcommand{\bs}{\boldsymbol{s}}
\newcommand{\bx}{\boldsymbol{x}}
\newcommand{\bphi}{\boldsymbol{\phi}}
\newcommand{\btheta}{\boldsymbol{\theta}}

\newcommand{\ftheta}{f_{\btheta}}

\newcommand{\Acal}{\mathcal{A}}
\newcommand{\Ncal}{\mathcal{N}}
\newcommand{\Scal}{\mathcal{S}}
\newcommand{\Tcal}{\mathcal{T}}

\newcommand{\norm}[1]{\left\lVert#1\right\rVert}

\DeclareMathOperator*{\argmax}{arg\,max}

\newcommand{\dint}[1]{\,\mathrm{d}#1}
\newcommand{\pder}[2]{\frac{\partial #1}{\partial #2}}

\newcommand{\dt}{\dint{t}}

\def\BibTeX{{\rm B\kern-.05em{\sc i\kern-.025em b}\kern-.08em
    T\kern-.1667em\lower.7ex\hbox{E}\kern-.125emX}}
\begin{document}

\title{
% Sample efficiency in model-based reinforcement learning
Learning a model is paramount for sample efficiency in reinforcement learning control of PDEs
}

\author{
\IEEEauthorblockN{2\textsuperscript{nd} Stefan Werner}
\IEEEauthorblockA{\textit{Department of Computer Science} \\
\textit{Paderborn University}\\
Paderborn, Germany \\
stefanwerner1997@gmail.com}
\and
\IEEEauthorblockN{1\textsuperscript{st} Sebastian Peitz}
\IEEEauthorblockA{\textit{Department of Computer Science} \\
\textit{Paderborn University}\\
Paderborn, Germany \\
sebastian.peitz@upb.de; ORCID: 0000-0002-3389-793X}
}

\author{Stefan~Werner, Sebastian~Peitz% <-this % stops a space
\thanks{Both authors are with the Department of Computer Science, Paderborn University, Paderborn, Germany. e-mail: sebastian.peitz@upb.de.}% <-this % stops a space
%\thanks{Manuscript received ??, ????; revised ??, ????.}
}

\maketitle

\begin{abstract}
The goal of this paper is to make a strong point for the usage of dynamical models when using reinforcement learning (RL) for feedback control of dynamical systems governed by partial differential equations (PDEs). To breach the gap between the immense promises we see in RL and the applicability in complex engineering systems, the main challenges are the massive requirements in terms of the training data, as well as the lack of performance guarantees. We present a solution for the first issue using a data-driven surrogate model in the form of a convolutional LSTM with actuation. We demonstrate that learning an actuated model in parallel to training the RL agent significantly reduces the total amount of required data sampled from the real system. Furthermore, we show that iteratively updating the model is of major importance to avoid biases in the RL training. Detailed ablation studies reveal the most important ingredients of the modeling process. We use the chaotic Kuramoto-Sivashinsky equation do demonstarte our findings. %As for the second problem, we discuss the implications of having a model -- possibly with prediction error bounds -- on the confidence of employing an RL agent in a real technical system.
\end{abstract}

\begin{IEEEkeywords}
reinforcement learning, surrogate modeling, partial differential equations, feedback control
\end{IEEEkeywords}

\section{Introduction}
Feedback control of complex physical systems is an essential building block in almost any modern technology such as wind energy \cite{aubrun2017review}, combustion \cite{wu2018jet} or nuclear fusion \cite{degrave2022magnetic}. In all these systems, we face the task of having to take control actions in a very short amount of time and for a system with highly complex, distributed dynamics (typically governed by nonlinear partial differential equations (PDEs)) that are difficult or even impossible to observe completely.
Since classical feedback control strategies have limitations for such highly complex systems, \emph{model predictive control} \cite{GP17} has emerged as an important control paradigm, where a model of the system dynamics is used to solve an optimal control problem repeatedly on a receding horizon in order to create a feedback loop. As this calls for the solution of an optimal control problem in a very short amount of time, surrogate models are frequently used to accelerate the solution process, in particular when the system dynamics are governed by PDEs. There exists a plethora of approaches, ranging from Galerkin models based on the \emph{Proper Orthogonal Decomposition} \cite{KV02} over the sparse identification of governing equations \cite{KKB18} to deep neural networks \cite{bieker2020deep}. All these modeling paradigms now face the challenge to trade between model accuracy and computational performance, which becomes more and more challenging for increasingly complex systems.

Due to the real-time challenges for MPC, \emph{reinforcement learning} \cite{sutton2018reinforcement} is gaining more and more popularity as a very powerful and real-time capable paradigm for feedback control. In recent years, we have seen tremendous successes, not only in the area of games (e.g., \cite{silver2016mastering}), but also in complex technical applications such as flow control \cite{rabault2019artificial} or nuclear fusion \cite{degrave2022magnetic}. However, there are two drawbacks that limit the deployment of RL agents in real systems.
% 1) the training often requires huge amounts of data and 
The first point is that the amount of required training data often exceeds tens to hundreds of millions of samples~\cite{mnih2015human,zoph2016neural} such that the training becomes very expensive. A notable example (even though from the area of game playing) includes the efforts of OpenAI Five to master the Dota 2 video game with an agent training on about 100,000 CPU cores for a time period of about ten months~\cite{berner2019dota}. These issues are even more severe when data collection becomes exceedingly resource-intensive, as is the case for most fluid or electromagnetic field simulations, see, e.g.,~\cite{ma2018fluid}.
The second drawback is that the control performance may vary strongly, in particular for high-dimensional state and action space dimensions, where it is challenging to obtain the required amounts of training data. 
A popular approach to tackle these issues is the usage of model-based algorithms \cite{sutton1991dyna}, see also \cite{wang2019benchmarking} for an overview and a taxonomy. Therein, a surrogate model replaces the real environment to allow for a significant increase of the training on data created by said surrogate. 
Using such a model enables the agent to look ahead at the effects of its actions on the system without ever performing them in the actual environment. 
It is widely accepted that surrogate models can reduce the amount of data an agent consumes before converging in the case that accurate and generalizable surrogates can be learned \cite{atkeson1997comparison,deisenroth2011pilco,moerland2020model},
and in many situations, this model can be orders of magnitude faster than evaluating the environment model (e.g., numerically solving a PDE).
Collecting data through \emph{model-based rollouts} is advantageous as soon as they reduce the amount of computation time spent on numerical simulations or the number of time-consuming and impractical real-world trials. 
At the same time, models introduce approximation errors so that we may obtain solutions inferior to those discovered using the model-free paradigm \cite{andersen1992evaluating}. The literature thus studies to what extent capable function approximators can overcome modeling bias and how to avoid for errors to propagate into control strategies \cite{chua2018deep,kurutach2018model,van2019use,janner2019trust}. Overall, interest has shifted from stable and robust linear models \cite{levine2014learning,lioutikov2014sample} to shallow neural networks \cite{xiong2014neuromechanical} or Gaussian processes \cite{seeger2004gaussian,deisenroth2011pilco} as learning machines, before advancing to the deep neural networks seen today in state-of-the-art algorithms for model-based control \cite{janner2019trust,hafner2019learning}. In the meantime, the increased model capacity makes overfitting a substantial issue of concern \cite{nagabandi2018neural,chua2018deep}. In spite of that, the recent \emph{SimPLe} \cite{kaiser2019model} and \emph{DREAMER} \cite{hafner2019dream} algorithms solve challenging visual control tasks, being orders of magnitude more efficient (even though the original problem is not expensive to simulate) than their model-free peers while rivaling their asymptotic performance.

% Similar to model predictive control, we now face the problem that careful modeling is required in order to avoid overfitting to an inaccurate model, which leads to biases in the learned control policy \cite{janner2019trust}.
% Alternatively, one can try to exploit system knowledge (in particular symmetries / invariances) in order to get smaller agents and thus to reduce the number of parameters that have to be trained \cite{BRV+19,PSC+23,GRS+23}.

Even though it seems clear that surrogates have the potential to improve the learning process, the trade off in terms of sample efficiency, computational complexity, and performance is seldom quantified, in particular in an engineering context such as PDE control. 
Two recent approaches using learned surrogates are \cite{liu2021physics} or \cite{zeng2022data}. 
% However, these do not provide a clear documentation of the required amount of data. 
In addition to assuming complete knowledge of the phenomena governing the system dynamics, the first work uses a physics-informed loss that, in essence, uses a deep neural network to compute the same finite difference solution as the numerical simulation does, but at a substantial slowdown in terms of runtime.
%The first work uses a physics-informed loss which in essence computes the same finite difference solutions as the numerical simulation does, which is a drawback in terms for efficiency. 
The second contribution learns a highly accurate surrogate before optimizing a control strategy on it but fails to adjust its forecasts to the continuously changing distribution of visited simulation states.
%In the second contribution, a global model with close to flawless forecasts has to be learned using random exploration.

%In this paper, we present an alternative surrogate modeling approach which is based on a convolutional autoencoder plus LSTM architecture (Section \ref{sec:learning-surrogates}). 
In this paper, we discuss and evaluate a variety of design choices associated with model-based reinforcement learning, adopt a more sophisticated optimization approach, and identify promising directions for future research.
We discuss the various modeling steps (Section \ref{sec:learning-surrogates}) that are required in order to arrive at a model that possesses the required accuracy.
%, and we also comment on the online updates necessary to avoid model overfitting. 
We then thoroughly study the impact on the required amount of training data (Section \ref{sec:learning-to-control-pdes}), where an improvement by a factor of nine is observed for the case of the 1D Kuramoto-Sivashinsky equation. 
In both Sections \ref{sec:learning-surrogates} and \ref{sec:learning-to-control-pdes}, we conduct various ablation studies to identify the most important modeling techniques in the offline phase, as well as online updating strategies during deployment.
Finally, we discuss the implications for the robustness of RL agents as well as future research directions in Section \ref{sec:discussion}.

\section{Reinforcement learning}
\label{sec:RL}
% We here only give a very brief overview of reinforcement learning (RL), a much more detailed introduction can be found in, e.g.,~\cite{sutton2018reinforcement}.
%The sequential control problem RL aims to solve is mathematically defined as a Markov Decision Processes (MDP). 
Reinforcement learning (RL) aims to solve sequential decision-making problems mathematically defined as Markov Decision Processes (MDPs) (see, e.g.,~\cite{sutton2018reinforcement} for a detailed overview).
It possesses a set $\Scal$ of system states and a set $\Acal$ of actions an \emph{agent} may select among to exercise control. At each discrete time step $\tau$, the agent observes state $\bs_{\tau} \in \Scal$ of the environment and responds with an action $\ba_{\tau} \in \Acal$. A stochastic transition function $\Tcal: \Scal \times \Acal \rightarrow \mathcal{P}(\Scal)$ formalizes in what way the system state changes as a result thereof. 
%In order for the next state $\bs_{\tau + 1}$ to be the result of a function with arguments $\bs_{\tau}$ and $\ba_{\tau}$, state transitions must be independent from past states and actions. In other words, 
In an MDP, the state transition is independent of past states and actions, meaning that it satisfies the \textit{Markov property}. The reward signal $r_{\tau}$ then quantifies the quality of decision $\ba_{\tau}$ taken in state $\bs_{\tau}$ and is an instance of the stochastic reward function $\mathcal{R}: \Scal \times \Acal  \rightarrow \mathcal{P}(\mathbb{R})$. Starting from an initial state $\bs_0 \sim \mathcal{P}_0(\Scal)$, the RL framework aims to maximize the expected sum of discounted future rewards $\mathbb{E}\left[\sum^{\infty}_{\tau=0} \gamma^{\tau}r_{\tau} \right]$, where $\gamma \in [0, 1]$ is the discount factor. % to account for potentially unbounded values in continuing tasks and weighs the contribution of future rewards to the overall objective value.

% \begin{figure}[b]
%     \centering
%     \includegraphics[width=\columnwidth]{figures/mdp.pdf}
%     \caption{The reinforcement learning control loop. An agent observes environment state $\bs_{\tau}$ and selects an action $\ba_{\tau}$ as its response, changing the environment to state $\bs_{\tau + 1}$ and returning reward $r_{\tau}$ as feedback. The Figure is adopted from~\cite{sutton2018reinforcement}.}
%     \label{fig:background:mdp-loop}
% \end{figure}

A policy $\pi$ is a mapping $\pi: \Scal \rightarrow \mathcal{P}(\Acal)$ modeling the probability $\pi(\ba | \bs)$ of taking action $\ba \in \Acal$ in state $\bs \in \Scal$ of the environment. Solving an MDP means finding an optimal policy $\pi^*$: % that yields at least the amount of future rewards in expectation as other policies:
\begin{align*}
    \pi^* &= \argmax\limits_{\pi} \mathbb{E}_{\pi}\left[\, \sum_{\tau = 0}^{\infty} \gamma^{\tau}r_{\tau}\,\right].
\end{align*}

So-called \textit{value functions} are useful concepts to formally define optimality conditions for policies. In substance, a \textit{state value} $V^{\pi}(\bs)$ denotes what future rewards to expect following policy~$\pi$ when the environment is in state $\bs$:
\begin{align*}
    %\label{eq:background:state-values}
    V^{\pi}(\bs) &= \mathbb{E}_{\pi}\left[\,\sum_{k=0}^{\infty}\gamma^{k}r_{\tau + k}   \biggm\vert \bs_{\tau} = \bs \,\right].
\end{align*}
Therefore, a policy $\pi$ is optimal if its state value $V^{\pi}\left(\bs\right)$ at each state $\bs \in \Scal$ is at least as large as the state values of any other policy. 
%Although multiple optimal policies $\pi^*$ may exist, optimal state values $V^*(\bs)$ are unique and shared among them. 
Similarly, \textit{state-action values} $Q^{\pi}(\bs, \ba)$ describe what rewards to expect once action $\ba$ is selected in state $\bs$ and policy $\pi$ governs the behavior thereafter:
\begin{align*}
    % \label{eq:background:state-action-values}
    Q^{\pi}\left(\bs, \ba\right) &= \mathbb{E}_{\pi} \left[ \, \sum_{k=0}^{\infty}\gamma^k r_{\tau + k}  \biggm\vert \bs_{\tau}=\bs, \ba_{\tau} = \ba  \, \right] \; .
\end{align*}

In \emph{Deep Reinforcement Learning}, deep neural networks are typically used as function approximators to learn policies and value functions.
%the policy $\pi$ (and in many cases the $Q$ function) are represented by deep neural networks. 
%, and then to search for the optimal policy based on rewards the agent receives upon interaction with the environment. 
For continuous control tasks (that is, continuous state and/or action spaces as in PDE control), \emph{policy gradient methods} (e.g., the \emph{Proximal Policy Optimization (PPO)} \cite{schulman2017proximal}, the \emph{deep deterministic policy gradient (DDPG)} \cite{lillicrap2015continuous} or the \emph{Soft Actor Critic (SAC)} \cite{haarnoja2018soft}) are the most prominent methods. %As the focus of this article is to emphasize the benefits of using models, we will not further go into detail regarding the specifics here.

\subsection{Model-based reinforcement learning}

Model-based algorithms use a \textit{dynamics model} to capture the changes in the environment, i.e., they approximate the state transition map $\Tcal: \Scal \times \Acal \rightarrow \mathcal{P}(\Scal)$ by a surrogate 
%$\ftheta: \Scal \times \Acal \rightarrow \mathcal{P}(\Scal)$ 
%of the decision-making process, 
$\ftheta\approx\Tcal$, where $\btheta$ are trainable parameters.
The model-based framework extends the optimization procedure of model-free reinforcement learning with additional \textit{planning} \cite{sutton2018reinforcement} and model-learning steps. 
%The design of dynamics models and their later purpose in improving control strategies sets model-based methods apart \cite{plaat2021high}. Therefore, t
%The literature distinguishes purely computational surrogates, simulating known state transitions in advance, from models learning the system dynamics in a data-driven manner \cite{moerland2020model}, see Figure \ref{fig:background:dyna-style-rl}. 
%Despite the recent successes of approaches developing plans for future actions with computational models, such as AlphaGo \cite{silver2016mastering}, % and AlphaZero \citep{silver2017mastering}, 
Our work focuses on models learned from data for two reasons. The phenomena defining the physics as well as their parameter values (for example mass or viscosity parameters) are often not known exactly. At the same time, computational models solving the governing PDEs often do not meet real-time constraints, which is essential for online planning.

\begin{figure}[h]
    \centering
    \includegraphics[width=\columnwidth]{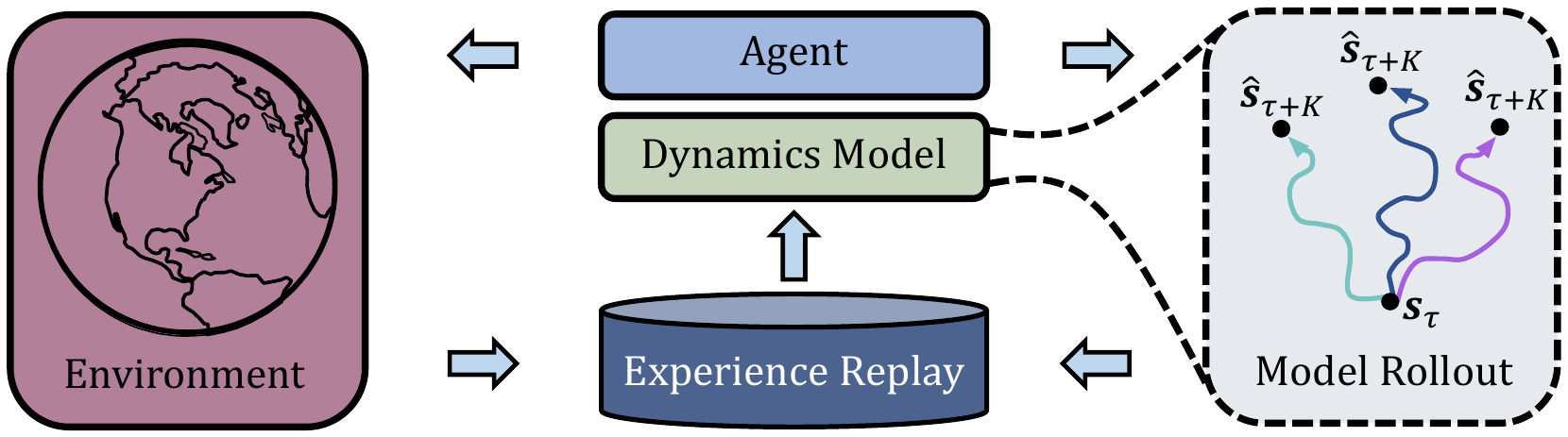}
    \caption{In model-based RL with learned models, collected data serves two purposes. Samples are used to learn a model of the system dynamics and to estimate update targets for the model-free agent. 
    %The dynamics model acts as a surrogate of the environment to the agent. 
    Using rollouts of the agent in the model, algorithms collect additional data to improve their behavior. In the ideal absence of prediction errors, the model matches the environment without its downsides of real-world trials or severe computational costs.}
    \label{fig:background:dyna-style-rl}
\end{figure}

%\textcolor{red}{To do: Longer discussion on modeling}

\subsection{Reinforcement learning for PDE control}

In the literature on partial differential equations, control problems are usually defined on a continuous time scale (in contrast to the discrete-time decision-making underpinning reinforcement learning). 
The system state $\bu:\Omega \times [0,T] \rightarrow \R^n$ is a function of time $t\in[0,T]$ and space $x\in\Omega$, and the dynamics is described by a nonlinear partial differential operator $\Ncal$, i.e., 
\[
	\pder{\bu}{t} = \Ncal(\bu,\bphi), 
\]
with $\bphi:\Omega \times [0,T] \rightarrow \R^m$ being the control input that may depend on both space and time. Moreover, the initial conditions are given by $\mathcal{I}(\bu,\nabla_{\bx} \bu, \ldots)$ and the boundary conditions by $\mathcal{B}(\bu,\nabla_{\bx} \bu, \ldots)$.

In order to draw the connection to reinforcement learning (or numerical implementations in general), one can introduce a partial discretization in time with a constant step size $\Delta \tau = t_{\tau + 1} - t_\tau$, $\tau=0,1,\ldots,p$, i.e.,
\[
	\Tcal(\bu_\tau, \bphi_\tau) = \bu_\tau + \int_{t_\tau}^{t_{\tau+1}} \Ncal(\bu(\cdot,t), \bphi_\tau) \dt = \bu_{\tau+1},
\]
where we assume that $\bphi(\cdot,t)=\bphi_\tau$ is constant over the interval $[t_\tau, t_{\tau+1})$.
Note that this time step $\Delta \tau$ is typically much larger than the time step $\Delta t$ in a standard numerical discretization scheme.
Using the above considerations, the control task can be formalized as an optimal control problem of the following form:
\begin{align}
    \min_{\bphi} J(\bu,\bphi) &= \min_{\bphi} \sum_{\tau=0}^{p} \ell\left(\bu_\tau,\bphi_\tau\right) \label{eq:OCP}\\
    \mbox{s.t.} \quad \bu_{\tau+1} &= \Tcal(\bu_\tau, \bphi_\tau), \qquad \tau = 0,1,2,\ldots,p-1, \notag
\end{align}
where $J$ is the objective functional over the time horizon $T = p\Delta \tau$, and $\ell$ is the \emph{stage cost}, e.g., a tracking term (with regularization, including penalties on the control cost) 
\begin{align*}
	\ell\left(\bu_\tau,\bphi_\tau\right) &= \norm{\bu_\tau - \bu_\tau^{\mathsf{ref}}}_{L^2}^2 + \lambda \norm{\bphi_\tau}_{L^2}^2.
\end{align*}
In terms of RL, the stage cost $\ell$ can be seen as the negative reward, the state $\bu$ corresponds to $\bs$ and $\bphi$ is linked to $\ba$.

\subsection{The Kuramoto-Sivashinsky equation}
The Kuramoto-Sivashinsky equation emerges in various reaction-diffusion systems and other physical phenomena, but was originally developed to model instabilities in laminar flame fronts. Since it admits chaotic behavior from a certain domain size on (and has similarities to the Navier-Stokes equations in fluid flow), it is one of the most frequently studied PDE systems in many situations.
The equation is defined as 
\begin{align}
    \label{eq:kuramoto-sivashinsky-equation}
    \frac{\partial \boldsymbol{u}}{\partial t} = - \nabla^2_{\boldsymbol{x}}\boldsymbol{u} - \nabla^4_{\boldsymbol{x}}\boldsymbol{u} - \frac{1}{2}\boldsymbol{u}\nabla_{\boldsymbol{x}}\boldsymbol{u} + \bphi \; ,
\end{align}
where $\boldsymbol{u}$ is the velocity and $\bphi$ is an additive forcing term. Similar to other works \cite{bucci2019control,zeng2022data}, we study a one-dimensional spatial domain $\Omega = [0, L]$ with periodic boundary conditions and system size $L = 22$. This spatial configuration has been shown to manifest stable chaotic conditions, having the smallest system size to develop sustained (weak) turbulence~\cite{cvitanovic2010state}. Again following related work \cite{bucci2019control,zeng2021symmetry,zeng2022data}, the control consists of a superposition of several Gaussians 
\begin{align}
\bphi_{x, \tau\Delta\tau} &= \sum^4_{i=1} \frac{\ba_{\tau}[i]}{\sqrt{2\pi\sigma}}\exp\bigg(-\frac{(\bx - \bx^{(i)})^2}{2\sigma^2}\bigg) \; ,
\end{align}
located at spatial coordinates~$\bx^{(i)} \in \{0, L/4, 2L/4, 3L/4\}$. 
Here $\ba_{\tau}[i] \in [-1, 1]$ is the $i$-th control output of the agent at time step $\tau$ and $\sigma = 0.4$. % defines the spread of the external force over the spatial domain. 
The applied control values $\ba_{\tau}[i]$ are changed at intervals of $\Delta\tau = 0.25$ time units. Each episode simulates $T_{\mathrm{max}} = 100$ time units of the system, that is, $400$ discrete steps, beginning with states sampled from the unforced attractor as an initial condition. In detail, we follow the configuration outlined in \cite{zeng2021symmetry} and initialize the solution variable $\boldsymbol{u}$ with uniform noise in the range $[-0.4,0.4]$. The simulation then continues developing the state evolution for $200$ time units (about 10 Lyapunov times). During the transient, the system is not actuated.

In accordance with \cite{zeng2022data,zeng2021symmetry}, our goal is to dampen the dissipation $D$ of the solution variable while minimizing the amount of energy $P $ spent to power the system as well as the actuation devices:
\begin{equation}
  \begin{split}
    D &= \langle \, \big[\nabla_{\boldsymbol{x}}^2 \boldsymbol{u}  \, \big]^2 \rangle \;\; \text{     and     }   
  \end{split}
  \begin{split}
    \;\; P &= \langle\, \big[\nabla_{\boldsymbol{x}} \boldsymbol{u}  \, \big]^2 \rangle + \langle \,  \boldsymbol{u}\bphi \, \rangle\; ,
  \end{split}
\end{equation}
where $\langle \, \star \, \rangle$ denotes the spatial average taken over the physical domain $\Omega$. The reward is thus 
\begin{equation}
    \label{eq:problems:reward-dissipation}
    r_{\tau} = - \int_{\tau \Delta\tau}^{\tau \Delta\tau + \Delta\tau} D(t) + P(t) \; \, dt \; .
\end{equation}

For the numerical solution, we implement the finite difference simulation outlined in \cite{liu2021physics} and use a fourth-order Runge-Kutta scheme with a fixed time step of $\Delta t = 0.001$. The second-order dissipation and fourth-order hyperviscous damping terms are discretized by a sixth-order central difference scheme, while a second-order upwind scheme approximates the convection term. The spatial domain is discretized into $N=64$ equidistant collocation points whose solution values constitute the state vector $\bs$ observed for decision-making. 

\section{Learning surrogate models of forced PDEs}
\label{sec:learning-surrogates}
Before integrating our surrogate into the online learning process of model-based RL, we evaluate its design on an offline dataset. In order to do so, we discuss the different modeling steps in detail and assess their importance in an ablation study on the Kuramoto-Sivashinsky system.

\subsection{Dimensionality reduction} 
\begin{figure}
    \centering
    \includegraphics[width=1.0\columnwidth]{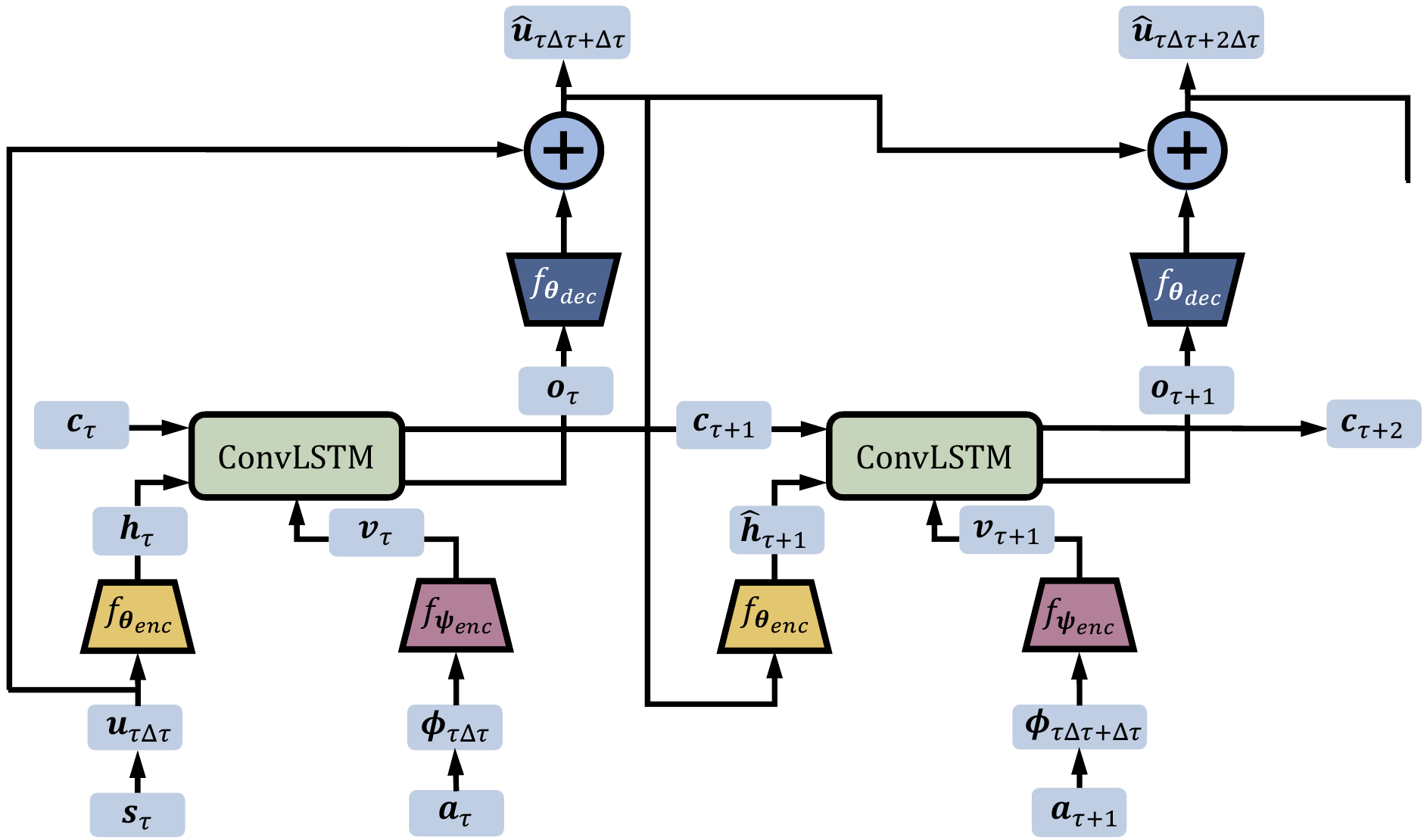}
    \caption{Our surrogate model is composed of a state encoder $f_{\boldsymbol{\theta}_{\mathrm{enc}}}$ (yellow), an action encoder $f_{\boldsymbol{\psi}_{\mathrm{enc}}}$ (red) and a convolutional LSTM cell $f_{\boldsymbol{\theta}_{\mathrm{fwd}}}$ (green) modeling the transition dynamics, as well as a decoder network $f_{\boldsymbol{\theta}_{\mathrm{dec}}}$ (blue) that restores the spatial extent of its output to the physical domain. Starting off an initial condition (left), the model predicts the state evolution using its prior output as an input of the state variable (right). The figure does not show intermediate scaling or normalization transformations for clarity.}
    \label{fig:learning-surrogates:autoreg-dynamics-model}
\end{figure}

For learning surrogates, the spatial extent of the physical domain $\Omega$ and the dimensionality of its discretization affects the amount of data necessary in training as well as the degree to which the surrogate matches the state evolution of the system. 
To achieve a dimensionality reduction, we use a \emph{convolutional autoencoder (CAE)} architecture, which is particularly well suited to cover the (on shorter time scales) local physics of PDEs. Another merit of CAEs is the weight-sharing capability, which significantly decreases the number of model parameters $\btheta$ and has a self-regularizing effect.
The architecture consists of an encoder and a decoder network, as illustrated by the yellow and dark blue blocks in Figure~\ref{fig:learning-surrogates:autoreg-dynamics-model}. The encoder network $f_{\boldsymbol{\theta}_{\mathrm{enc}}}$ compresses snapshots $\bs_{\tau}$ of the system to a compact latent space $\boldsymbol{h}_{\tau} = f_{\boldsymbol{\theta}_{\mathrm{enc}}}(\boldsymbol{s_{\tau}})$, while the decoder network $f_{\boldsymbol{\theta}_{\mathrm{dec}}}$ attempts to recover the original input $\hat{\bs}_{\tau} =  f_{\boldsymbol{\theta}_{\mathrm{dec}}}(\boldsymbol{h}_{\tau})$. The information bottleneck between the encoder and decoder networks regulates the degree to which the state dimensionality is reduced. 
Supposing that the reconstruction loss $\mathcal{L}_{\mathrm{MSE}}$ is small, the latent variable $\boldsymbol{h}_{\tau}$ holds similar information as the original input~$\bs_{\tau}$ but in denser form. More details on the specifics of the network architecture can be found in Appendix \ref{app:NN}.

\subsection{Time stepping}
To take full advantage of the fact that both the latent state $\boldsymbol{h}_{\tau}$ and the auto-encoded control $\boldsymbol{v}_{\tau}$ (Appendix \ref{app:NN}) act in the same spatial domain, albeit downsampled, we use a convolutional LSTM model \cite{shi2015convolutional} to transition in latent space.
Combining all components outlined so far, we pass solution variables $\boldsymbol{u}_{\,\tau\Delta\tau}$ and forcing terms $\bphi_{\,\tau\Delta\tau}$ to the encoder networks $f_{\boldsymbol{\theta}_{\mathrm{enc}}}$ and $f_{\boldsymbol{\psi}_{\mathrm{enc}}}$ and compute $\boldsymbol{h}_{\tau} = f_{\boldsymbol{\theta}_{ \mathrm{enc}}}(\boldsymbol{u}_{\,\tau\Delta\tau})$ and $\boldsymbol{v}_{\tau} = f_{\psi_{ \mathrm{enc}}}(\bphi_{\, \tau\Delta\tau})$ before transitioning in latent space with $f_{\boldsymbol{\theta}_{\mathrm{fwd}}}$ to predict the deterministic state evolution of the decision-making process. Similar to other works \cite{mo2019deep,geneva2020modeling,ren2022phycrnet}, the decoder network $f_{\boldsymbol{\theta}_{\mathrm{dec}}}$ is tied into a network learning temporal residuals, i.e., it is 
used to predict state changes $\Delta\boldsymbol{s}_{\tau} = \boldsymbol{s}_{\tau + 1} - \boldsymbol{s}_{\tau}$ of solution variables. State changes are scaled with the time duration $\Delta \tau$ of control steps and modeled in latent space. Therefore, our network architecture infers state transitions as
\begin{align}
    \boldsymbol{\hat{u}}_{\,\tau\Delta\tau + \Delta\tau} &= \boldsymbol{u}_{\,\tau\Delta\tau} + \Delta\tau \cdot f_{\boldsymbol{\theta}_{\mathrm{dec}}}(\boldsymbol{o}_{\tau}) \; ,\\
    \text{with } \;\; \boldsymbol{o}_{\tau} &= f_{\boldsymbol{\theta}_{\mathrm{fwd}}}(\boldsymbol{h}_{\tau}, \boldsymbol{v}_{\tau}; \boldsymbol{c}_{\tau}) \; ,
\end{align}
where targets $\Delta\boldsymbol{s}_{\tau} / \Delta \tau \approx f_{\boldsymbol{\theta}_{\mathrm{dec}}}(\boldsymbol{o}_{\tau})$ are used to train the transition model $f_{\boldsymbol{\theta}_{\mathrm{fwd}}}$ via end-to-end backpropagation. The variable $\boldsymbol{c}_{\tau}$ is the cell state of our recurrent transition model. The overall network architecture is illustrated in Figure~\ref{fig:learning-surrogates:autoreg-dynamics-model}. It shows that, in each step, our model 
(1) transforms control outputs $\boldsymbol{a}$ to external forcing terms $\bphi$ using a known functional,
(2) encodes solution variables $\boldsymbol{u}$ and forcing terms $\bphi$ to a latent space defined on a shared spatial domain with encoder neural networks (yellow and red components),
(3) uses a transition model (green component) to infer encoded state changes $\boldsymbol{o}$, and 
(4) decodes temporal changes $\boldsymbol{o}$ (blue component) before adding them to the original input.
In order to use our model as a surrogate of the system, it is unrolled in time, taking the control output of an agent as well as its prior prediction for an input.

A notable benefit of learning temporal residuals is that the network's predictions obey the initial conditions of the system by construction. In fact, modeling state changes scaled with coefficient $\Delta\tau$ imposes a temporal smoothness assumption on the predictions to thwart irregular and erratic changes in the state evolution. In the context of model-based control, our temporal smoothness assumption offers a significant advantage. Since the control loop intertwines data collection, model learning, and behavior improvement, only a limited number of system snapshots are available to the surrogate at first. As we confirmed in our experiments, models predicting states in place of state changes often suffer from compounding model errors in the small data regime due to their dependence on well-functioning encoder and decoder networks. In the small data regime, said neural networks are often still unable to encode and decode input states to a sufficient degree of accuracy, which can result in large model errors that can cause \textit{catastrophic learning updates} \cite{van2019use} for the agent. In contrast, the additive changes of our model architecture couple its predictions to the initial condition and enforce a close proximity to the temporal state evolution of the system for at least a few steps of the prediction until model errors begin compounding. 

\section{Learning to control PDEs sample efficiently}
\label{sec:learning-to-control-pdes}

As a member of the \emph{Dyna} family of model-based algorithms \cite{sutton1990integrated,sutton1991dyna}, our approach implements a model-free agent to learn behavior $\pi_{\boldsymbol{\omega}}$ and improve it over time. 
In essence, all descendants of Dyna follow instructions similar to those outlined in Alg.~\ref{alg:controlling:vanilla-mbrl-algorithm}.
In a broad sense, the algorithm intertwines data collection, model learning, and behavior improvement for its main steps, as shown in Alg.~\ref{alg:controlling:vanilla-mbrl-algorithm}. Each iteration begins with rolling out $\pi_{\boldsymbol{\omega}}$ to collect additional samples that are later stored as a dataset $\mathcal{D}_{\mathrm{env}}$. After training a deep neural network $f_{\boldsymbol{\theta}}$ to learn the dynamics of the environment on samples stored in $\mathcal{D}_{\mathrm{env}}$, the model serves as an approximate MDP $\mathcal{M}^\prime$ of the actual system $\mathcal{M}$. 
Supplementary samples from model-based rollouts of $\pi_{\boldsymbol{\omega}}$ are then stored in a separate dataset $\mathcal{D}_{\mathrm{model}}$ as proxies for the actual experience. Iterations of our algorithm conclude with an update of the model-free agent. Here, data stored in $\mathcal{D}_{\mathrm{env}}$ as well as $\mathcal{D}_{\mathrm{model}}$ is used, and the latter amount often exceeds the other by multiple orders of magnitude.

\begin{algorithm}[t]
	\caption{Model-Based Reinforcement Learning}\label{alg:controlling:vanilla-mbrl-algorithm}
\begin{algorithmic}[1]
	\State Initialize parameters of policy $\pi_{\boldsymbol{\omega}}$ and model $f_{\boldsymbol{\theta}}$\;
	\State Initialize empty datasets $\mathcal{D}_{\mathrm{env}}$ and $\mathcal{D}_{\mathrm{model}}$\;
	\While{not done}
	\State Sample $\mathcal{M}$ using policy $\pi_{\boldsymbol{\omega}}$ \hfill$\rightarrow$ add to $\mathcal{D}_{\mathrm{env}}$\;
	\State Train model $f_{\boldsymbol{\theta}}$ on dataset $\mathcal{D}_{\mathrm{env}}$\;
	\State Sample $\mathcal{M}^{\prime}$ using policy $\pi_{\boldsymbol{\omega}}$ \hfill$\rightarrow$ add to $\mathcal{D}_{\mathrm{model}}$\;
	\State Update policy $\pi_{\boldsymbol{\omega}}$ using samples of $\mathcal{D}_{\mathrm{model}}$ and $\mathcal{D}_{\mathrm{env}}$\;
	\EndWhile
\end{algorithmic}
\end{algorithm}

\subsection{The model-based RL framework}
A concrete implementation of the model-based principle is, therefore, all about 
(1) the design of a surrogate and its training process,
(2) the way in which $\mathcal{D}_{\mathrm{model}}$ is populated with artificial samples, and
(3) the model-free agent updating~$\pi_{\boldsymbol{\omega}}$ with data stored in $\mathcal{D}_{\mathrm{env}}$ and fictitious samples in $\mathcal{D}_{\mathrm{model}}$.
In the following, we outline our approach to each of the above steps for model-based reinforcement learning. 

\subsubsection{Model learning} 
Unlike related work on model-based fluid control, we do not collect snapshots of the system in advance using random exploration. Instead, we alternate between data collection and model learning to align the distribution of states that PI visits with the data our surrogate is trained on and approximates well.
Our experiments (Section \ref{sec:numerics_RL}) suggest that an online adaptation of the model to changes in the behavior is essential for the model to maintain its relevance.
%Snapshots and state transitions of the system are not given in advance but are collected gradually with each iteration, starting with a random exploration policy~$\pi_{\mathrm{rand}}$.
%Unlike random exploration, using the learned $\pi_{\boldsymbol{\omega}}$ in later iterations aligns the distribution of visited states with whatever behavior the agent deems most promising. 
%This online adaptation of the model to changes in the behavior is an essential component of our algorithm.
%
After each step of data collection, We train the model $f_{\boldsymbol{\theta}}$ using the samples collected in $\mathcal{D}_{\mathrm{env}}$. To mitigate overfitting in the small data regime, we monitor the model on a validation set $\mathcal{D}_{\mathrm{val}}$ and stop training early once the validation loss $\mathcal{L}_{\mathrm{val}}$ converges. 
%As is standard practice~(see, e.g.,~\citep{nagabandi2018neural}), we hold out sequences of entire episodes in place of single-step transitions from dataset~$\mathcal{D}_{\mathrm{env}}$ to estimate the validation loss. Otherwise, the validation loss does not estimate the ability of our model to generalize its forecasts to unseen actuation patterns of the agent. 
% After the initial stage of learning, the model should approximate the system dynamics fairly well on regions of the state space already covered during the initial exploration. Later iterations merely tweak the model to the shifting state distribution given a growing number of samples. In line with this intuition, 
% We steadily reduce the number~$P_{\mathrm{val}}$ of validation checks without improvement before stopping the training (the \textit{patience} parameter of early stopping). In this way, the model is first trained until convergence and afterward updated using fewer gradient steps. 
% Other works (see, e.g.,~\citep{kaiser2019model}) define a fixed number of training steps per iteration instead, which we found to be inflexible and prone to under- or overfitting. After the initial training period, the model is updated in regular intervals of $L_{\mathrm{model}}$ iterations given an additional number of $N_{\mathrm{iter}}$ samples per intermediate rollout of $\pi_{\bphi}$ on the environment. The optimizer preserves its state (for example the adaptive learning rate or momentum) throughout iterations but resets the violation count of validation checks before each update. 
In order to encourage an adaptation of the model to recent experience but protect~$\pi_{\boldsymbol{\omega}}$ against severe changes of inputs during policy training, we limit the minimum $P_{\mathrm{min}}$ and maximum $P_{\mathrm{max}}$ number of gradient updates to the model. We found that striking a balance between regular model updates but keeping shifts in the distribution of artificial samples minor helped stabilize the optimization. In our experiments, continuous updates of the model each performing only a few gradient steps (small values for $P_{\mathrm{max}}$ and $P_{\mathrm{val}}$) generally worked well and outperformed updating the model for an unlimited number of steps.
We also found that using a curriculum (similar to \cite{chiappa2017recurrent}) for the data collection is beneficial. We use short model-based rollouts of policy $\pi_{\boldsymbol{\omega}}$ for artificial data collection early on, and increase their prediction horizon once more data is available \cite{janner2019trust}. In other words, we optimize the model for short-term performance in the small data regime and emphasize learning global dynamics later on.

\subsubsection{Model-Based Rollouts}
Consistent with standard practice \cite{kurutach2018model,chua2018deep}, we mitigate the risk of model exploitation using an ensemble $\{f_{\boldsymbol{\theta}_1}, \dots, f_{\boldsymbol{\theta}_{L_{\mathrm{ens}}}}\}$ of dynamics models. 
The models deviate not only by their initial weights and the ordering of mini-batches, but also in terms of the data used for training and validation. 
Indeed, using a bootstrapping mechanism for training, \cite{chua2018deep} find that ensembles isolate epistemic uncertainty through disagreement. Since epistemic uncertainty is due to data scarcity, ensembling helps identifying regions of the state space where overfitting is likely and forecast errors are substantial.  
As is standard practice (see, e.g.,~\cite{janner2019trust}), we make use of the \textit{elite mechanism} for the ensemble. More details on our ensemble techniques can be found in Appendix \ref{app:ensembles}.

Using our ensemble of dynamics models $\{f_{\boldsymbol{\theta}_1}, \dots, f_{\boldsymbol{\theta}_{L_{\mathrm{ens}}}}\}$, we define an approximate MDP $\mathcal{M}^{\prime}$ imitating the original system $\mathcal{M}$. %Collecting data through model-based rollouts is advantageous in case they reduce the amount of computation time spent on numerical simulations or the number of time-consuming and impractical real-world trials. 
Analogous to recent works on Dyna algorithms, model-based rollouts in our implementation branch off arbitrary system states sampled from dataset $\mathcal{D}_{\mathrm{env}}$. Unlike rollouts beginning at initial conditions of the decision-making process, branching rollouts off arbitrary starting states avoids compounding model errors for states visited during later stages of episodes. In place of sampling single starting states, we select state-action sequences $(\boldsymbol{s}_{\tau - K_{\mathrm{tf}}}, \boldsymbol{a}_{\tau - K_{\mathrm{tf}}}, \dots, \boldsymbol{s}_{\tau}, \boldsymbol{a}_{\tau})\sim \mathcal{D}_{\mathrm{env}}$ of length $K_{\mathrm{tf}}$. Sampling sequences enables us to warm-start the memory unit of our recurrent transition model $f_{\boldsymbol{\theta}}$. Each state-action pair of the sequence is then processed in teacher-forcing mode to guide the transitions. We match the sequence length $K_{\mathrm{tf}}$ to the number of teacher-forcing steps during training. In our experiments, using~$K_{\mathrm{tf}}$ state transitions to seed the cell state was essential since we did not backpropagate gradients to its initial state at training time. Vice versa, we do not learn the initial condition of the memory unit since model-based rollouts start in states at different points in time for the system, which otherwise destabilized the algorithm in our experiments. In order to implement a similar mechanism for the first $K_{\mathrm{tf}}$ steps of an episode, we use a simple padding method and repeat the initial state-action tuple to extend the sequence to length~$K_{\mathrm{tf}}$. 
%In future work, transient steps of the simulation could serve the same purpose, that is, states of the temporal evolution before the controller applies external forcing.
A longer discussion on our rollout approach has been shifted to Appendix \ref{app:rollouts}.

\subsubsection{Policy optimization}

As our approach belongs to the Dyna family, arbitrary model-free agents can be used to derive behavior given artificial samples. An important design choice for our purpose is whether to optimize the behavior using on-policy or off-policy methods. Indeed, members of the former category discard existing experience after each update to the policy. Despite our dynamics models being designed as lightweight surrogates of the numerical simulation, reusing already existing experience is still more efficient. Adopting the soft-actor critic \cite{haarnoja2018soft} as our off-policy method (as in \emph{MBPO} \cite{janner2019trust}) the samples our model synthesizes in each iteration can be kept in dataset $\mathcal{D}_{\mathrm{model}}$ for a number of repetitions. Using artificial samples, the model-based algorithm updates the soft-actor critic $L_{\mathrm{updates}}$ times per iteration. Unlike the model-free framework, parameter $L_{\mathrm{updates}}$ is set in such a way that multiple updates (often between 20 and 40, see, e.g., \cite{janner2019trust}) are performed per environment interaction. To ensure the stability of the agent even after updating the behavior more than once, we collect a large number of artificial samples $N_{\mathrm{model}}$ in each iteration (in \cite{janner2019trust}, for example, 400 model rollouts are performed for each sample of the environment). After model-based data collection, the agent uses the samples stored in $\mathcal{D}_{\mathrm{env}}$ and $\mathcal{D}_{\mathrm{model}}$ to update its state-action value estimates and policy.

% \subsubsection{Implementation details} 
Our work builds on code made openly available in \cite{Pranjal2018} and \cite{yarats2021github}. The soft-actor critic agent implements double Q-learning\cite {van2016deep} to learn estimates of state-action values with separate networks $Q_{\boldsymbol{\theta}}$ and $Q_{\boldsymbol{\theta}^{\prime}}$ to mitigate selection biases and uses delayed target networks $Q_{\boldsymbol{\theta}^{-}}$ to stabilize the optimization towards a moving target. Soft updates of $Q_{\boldsymbol{\theta}^{-}}$ through Polyak averaging \cite{polyak1992acceleration} keep the target parameters consistent \cite{mnih2015human}. Similar to the TD3 algorithm \cite{fujimoto2018addressing}, multiple gradient updates of the critic are taken before altering the actor to ensure up-to-date value estimates. The actor $\pi_{\boldsymbol{\omega}}$ parameterizes the mean and covariance of a Gaussian distribution defined over the continuous action space. Our implementation does not use automatic entropy tuning \cite{haarnoja2018soft} but sets the temperature to a constant value instead. Both the actor and critic are implemented as multilayer perceptrons of three layers with $256$ hidden units each.

\section{Experimental Evaluation}
\label{sec:numerics}
Corresponding to Sections \ref{sec:learning-surrogates} and \ref{sec:learning-to-control-pdes}, we first evaluate the capabilities of our surrogate to approximate the state evolution well when the data is collected in advance, before considering the control aspect. More details on the training of our model can be found in Appendix \ref{app:training}.

\subsection{Data-driven prediction of forced PDEs}

%We here report the numerical results for the prediction model. 

% \subsubsection{Data collection}
% We collect snapshots of a system under an external force with control outputs sampled uniformly at random. We systematically change the number of episodes making up the training dataset to evaluate how different amounts of data affect performance. In line with our later use of surrogates in model-based reinforcement learning, we evaluate its predictive capabilities for sequences of length $K_{\mathrm{max}} + K_{\mathrm{tf}}$ subsampled from the testing episodes. 

% Each episode simulates 400 state transitions spanning $T_{\mathrm{max}} = 100$ time units where actions are changed every $\Delta\tau=0.25$ time units. We split the samples into training, validation (80\% / 20\%), and testing episodes. All results are the aggregate of cross-validation scores with five folds. 
% We study the evolution of state variable $\boldsymbol{u}$ for subsequences with $K_{\mathrm{max}} = 30 $ consecutive prediction steps ($7.5$ time units) during training and testing. 

In order to evaluate the amount of data our surrogate needs to approximate the state evolution, we collect a dataset of $100$ simulation episodes ($40,000$ snapshots), where actions are sampled uniform at random. We split the samples into training,
validation (80\% / 20\%), and testing episodes. All results are the aggregate of cross-validation scores with five folds. In general, we study the evolution of state variable $\boldsymbol{u}$ for subsequences with $K_{\mathrm{max}} = 30 $ consecutive prediction steps ($7.5$ time units) during training and testing. 

\begin{figure}[b]
    \includegraphics[width=1.0\columnwidth]{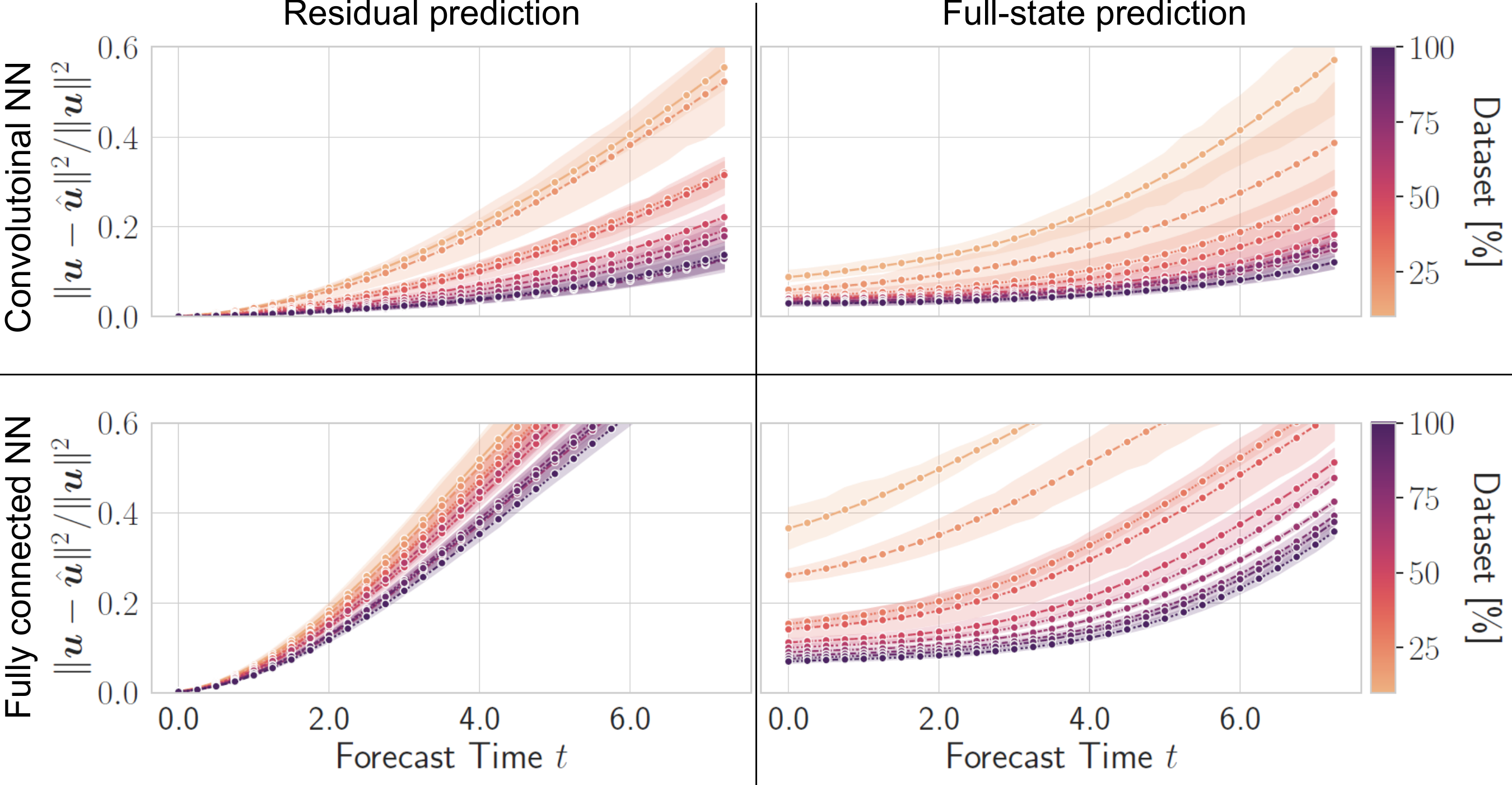}
    \caption{Prediction error of the Kuramoto-Sivashinsky equation. Each datapoint averages the error for different testing subsequences. The models are trained on an increasing share of the dataset composed of $40,000$ snapshots. Solid lines depict the mean of five folds, while the shaded regions show the 95\% confidence interval.}
    \label{fig:learning-surrogates:uvar-plot}
\end{figure}

Since our later approach to model-based approach aims to use as few samples as possible, the small data regime is critical to our work. To assess the alignment of our model design with our overall goal, we systematically decrease the amount of data available throughout cross-validation from $100\%$ down to $10\%$. 
%Since our later approach to model-based RL learns surrogates and control strategies using as few interactions with the environment as possible, the small data regime is critical to our work. To assess the accuracy of models in the small and large data regime, we gradually decrease the amount of data given for cross-validation from $100\%$ down to $10\%$ of the original $100$ episodes, that is, $40,000$ snapshots. 
In fact, our dataset (including validation and testing data) equals the number of snapshots used in the work of \cite{zeng2022data} for modeling the Kuramoto-Sivashinsky equation upfront before they derive a control strategy with the surrogate afterward. 
%Our work later shows the benefits of alternating model learning and policy improvement, finding reasonable control strategies in a fraction of the amount. 
In order to isolate the contribution of single design decisions for our dynamics model, we introduce several ablations and compare our results with those of a simple baseline, namely (1) the usage of fully connected networks instead of CNNs and (2) the prediction of the state instead of a residual update.
We train each model for a maximum of $250$ epochs or terminate the optimization once the monitored loss does not decrease for $P_{\mathrm{val}}= 25$ validation epochs. 

\begin{figure}
    \centering
    \includegraphics[width=1.0\columnwidth]{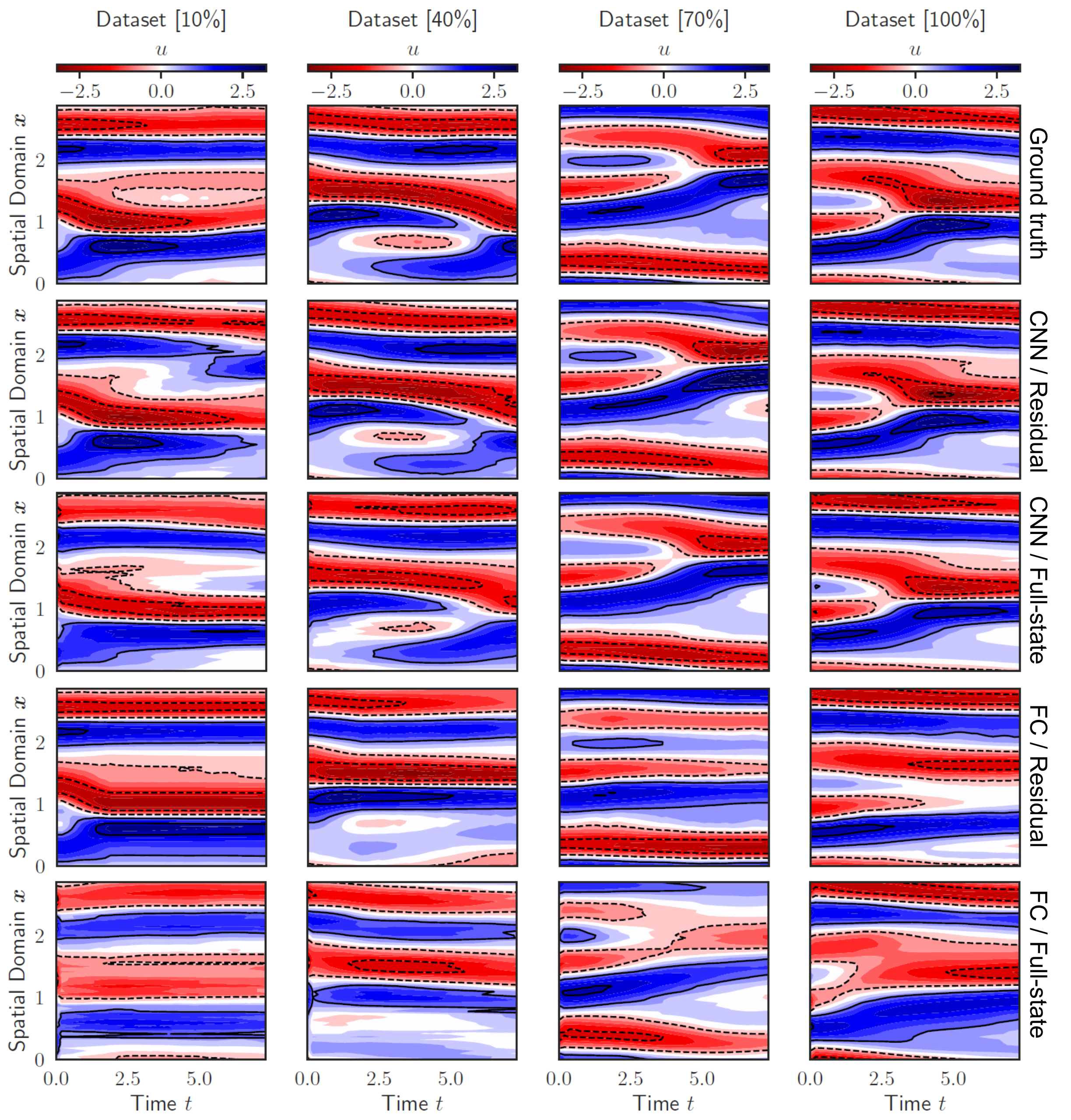}
    \caption{Comparison of the temporal state evolution of the Kuramoto-Sivashinsky equation with forecasts of our model and the different ablations. }
    \label{fig:learning-surrogates:visualization}
\end{figure}

In Figure~\ref{fig:learning-surrogates:uvar-plot}, we show the normalized mean squared state prediction error, averaged over various initial conditions. Unsurprisingly, convolutional models (top row) prove far more effective than fully connected networks (bottom row) in both the small as well as large data regime, learning temporal residuals or not. In terms of the comparison between residual and full-state prediction (left vs.\ right column), the former yields a notable advantage for forecasts over short time durations with error margins being negligible for several time units. In the small data regime, the full-state prediction is not yet equipped to recover state variables and, therefore, suffers from large error margins. A notable advantage of our approach is thus the strongly reduced data requirements supposing we limit the prediction horizon appropriately. Since our RL approach will limit model usage after a small number of prediction steps, the inferior accuracy after $\sim3.25$ time units (13 discrete steps) is of no concern to the optimization. The errors in the predicted rewards (which have a significant impact on the learning performance) are very similar to the state prediction. An example of forecasts for each model at an increasing share of the $40,000$ snapshots for training is shown in Figure~\ref{fig:learning-surrogates:visualization}. 

In general, increasing the degrees of freedom of a model enables it to learn the state evolution of snapshots in greater detail. At the same time, models of higher capacity are at a greater risk of overfitting to patterns in the training set, contriving substantial errors once presented with data withheld for testing. Considering this trade-off is essential for model-based RL since overfitting is among its greatest challenges \cite{chua2018deep,nagabandi2018neural,kurutach2018model}. A study on this dependency can be found in Appendix \ref{app:NNsize}. For the main body of this paper, all models have the identical structure, which we found to be a good trade off between expressive capabilities and complexity.

\subsection{Reinforcement learning control of PDEs}
\label{sec:numerics_RL}
At last, we evaluate the ability of our approach to model-based reinforcement learning to stabilize systems with dynamics governed by PDEs. In this section, we study whether 
(1) learning a surrogate of the global dynamics indeed mitigates the data consumption of model-free approaches,
(2) learned control strategies can match their effectiveness, although model-based reinforcement learning algorithms introduce approximation errors to the experience used for behavior improvement, 
(3) examine the contribution of single components to our overall approach in isolation, and
(4) evaluate our algorithm for a number of configurations.

\subsubsection{Data efficiency of learned control laws}

To assess the capabilities of our control algorithm in mitigating the data consumption of the model-free RL framework, we compare accomplished episode returns to those of the popular PPO \cite{schulman2017proximal} and SAC \cite{haarnoja2018soft} algorithms. As outlined before, the former algorithm belongs to the family of controllers most often encountered in the fluid mechanics literature nowadays (see, e.g., \cite{viquerat2021review} for a comprehensive survey), despite the agent discarding past experiences after each update. Since our approach integrates SAC as a main component for policy improvement, it is the candidate best suited to compare against. In both cases, we use the open-source implementations made available by the \textit{stable-baselines3} project \cite{stable-baselines3} as well as their default configuration for our experiments.

\begin{figure}[bh]
    \centering
    \includegraphics[width=1.0\columnwidth]{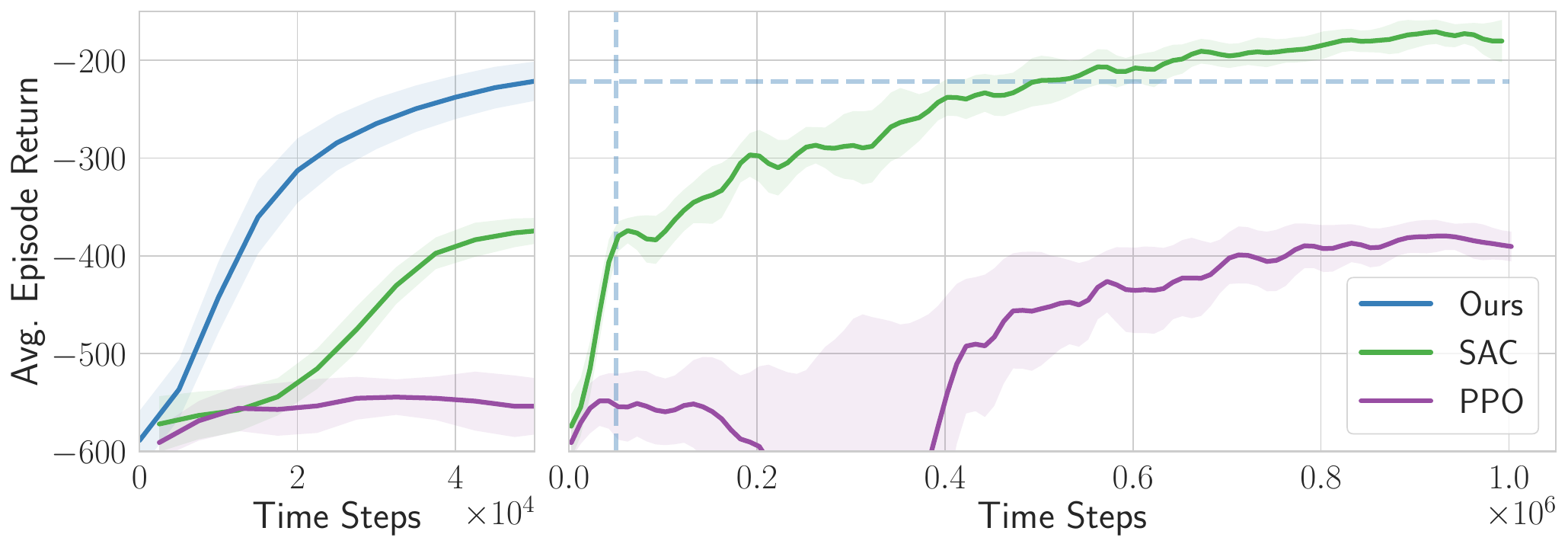}
    \caption{Average evaluation episode returns for our approach compared to SAC and PPO for up to $50,000$ (left) and $1,000,000$ (right) steps of training. Solid lines show the mean of three trials, while shaded regions denote the standard deviation among trials. All lines are smoothed using a Gaussian filter with $\sigma = 1$. The dotted lines on the right-hand side show the average performance of our model-based algorithm after termination at $50,000$ time steps.}
    \label{fig:learning-control:mbrl-ppo-sac-comparison}
\end{figure}

\begin{figure}[t]
    \centering
    \includegraphics[width=1.0\columnwidth]{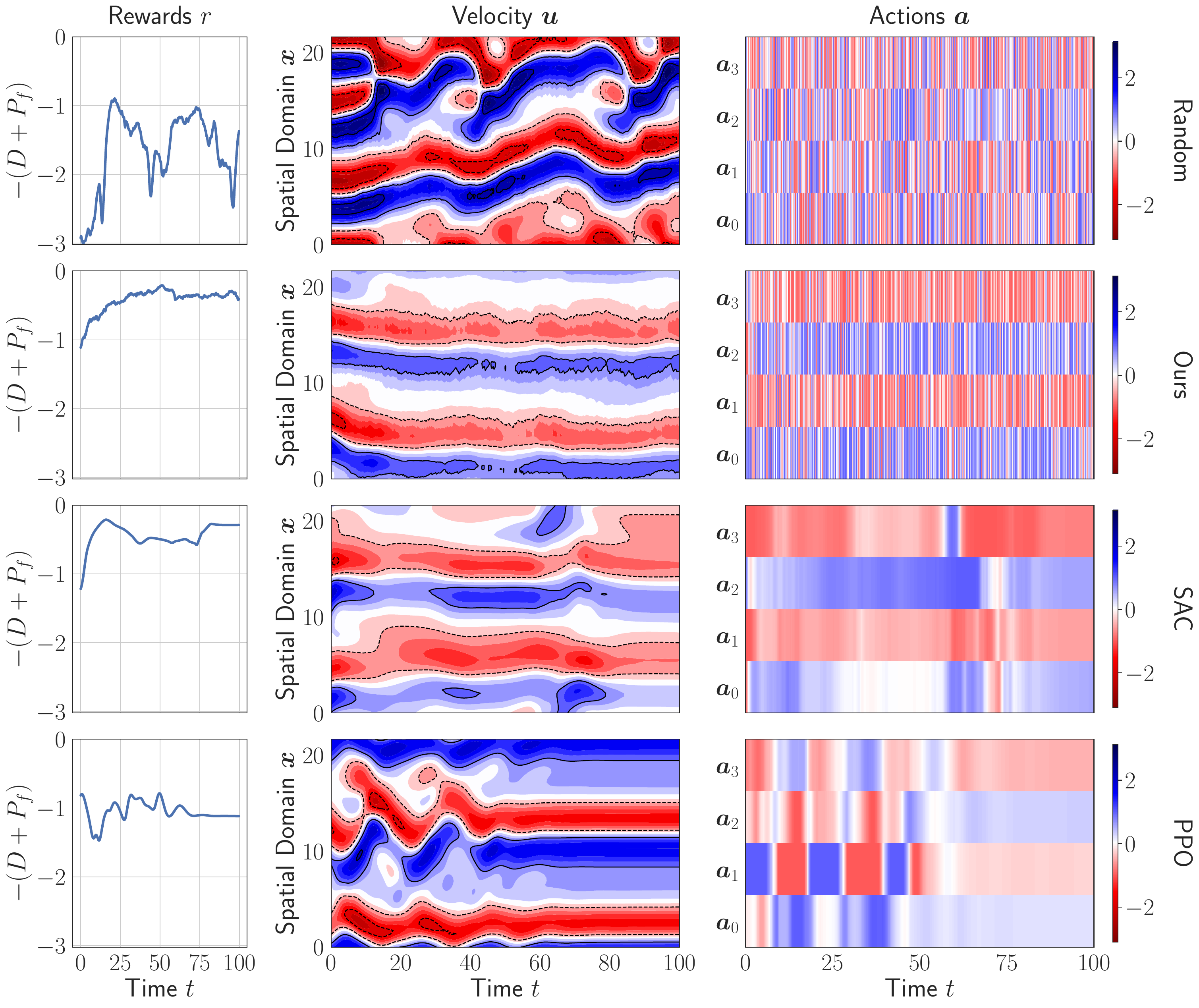}
    \caption{The spatio-temporal evolution of the Kuramoto-Sivashinsky system (middle) and its dissipative term $D$ and power consumption $P$ (left) as a result of the actions applied to it (right). In each row, we show an agent attempting to stabilize the dynamics starting from an initial condition sampled at random. The figure shows control strategies after their training terminated after $50,000$ or $1,000,000$ steps, respectively.}
    \label{fig:learning-control:mbrl-show-control-strategies}
\end{figure}

Figure \ref{fig:learning-control:mbrl-ppo-sac-comparison} shows the average return of evaluation episodes with respect to the number of samples collected for training thus far. Due to computational costs, we terminate the execution of our model-based approach once it obtained $50,000$ samples. Since computational workloads per interaction with the system are lower for model-free algorithms, we execute our baselines for $1,000,000$ samples. On the left-hand side of the figure, we show the training curves up to the point where we terminate our algorithm, while the right-hand side shows the asymptotic behavior of  SAC and  PPO. We find that our model-based approach indeed learns suitable control strategies using substantially fewer samples. On average, SAC takes about $430,000$ samples to break even with the average episode returns of our approach at $50,000$ samples, while PPO cannot match its performance even after $1,000,000$ steps. With an almost nine-fold improvement in terms of data consumption over SAC, this demonstrates that model-based RL is indeed a promising direction to learn control strategies for systems governed by PDEs. %Unfortunately, however, we were unable to conduct an evaluation of the asymptotic behavior for our model-based approach due to computational limitations. Still, its
Even though we stopped training early after $50,000$ steps, the control strategy dampened the dissipation $D$ and power consumption $P$ of the system by more than $63\%$ of the values accomplished with random forcing. At the same time, SAC converges to a value of about $73\%$, although using $20$ times the amount of data to do so. 
%
% Overall, our model-based approach indeed manages to establish control strategies capable of stabilizing the chaotic regime after brief time periods. 
An example of what actuations the controller applies to navigate the system towards a stable state after termination is illustrated in Figure~\ref{fig:learning-control:mbrl-show-control-strategies}. %The figure shows evaluation episodes beginning in distinct initial conditions and includes examples of the temporal state evolution under random forcing and stabilized with the SAC and PPO agents for reference.

\subsubsection{Ablation study}

A part of the motivation for our work are the shortcomings of past studies on model-based control for governed systems insofar as recent advances in the reinforcement learning literature are not properly taken into account. As outlined before, the related works \cite{liu2021physics} and \cite{zeng2022data}, for example, use neither ensembling nor rollout scheduling and in the latter case do not adapt the model to altering state distributions, despite each feature being considered standard practice in other fields. In order to appraise the contribution of each component to the overall performance of our method in isolation, our work examines the following ablations of our model-based approach.

\begin{itemize}
    \item \emph{Offline model training ablation:} Similar to \cite{zeng2022data}, we collect 50,000 snapshots of the system in advance using random exploration. After training our ensemble to mimic the global dynamics, the RL agent acts entirely within the surrogate for data collection. %Other than \citeauthor{zeng2022data}~\citep{zeng2022data}, model rollouts branch off states stored in experience replay memory in place of the initial conditions of the system, which we found to be essential for mitigating compounding model errors. Should our ablation succeed in learning suitable control strategies, it would suggest that random exploration of the state space is sufficient to learn about the global dynamics of the system, that is, to approximate the system when it approaches an equilibrium, despite the model being trained in the chaotic regime. The ablation performs the same number of gradient updates to the model-free agent as our approach.
    \item \emph{Model exploitation ablation:} Our approach establishes multiple mechanisms to prevent compounding model errors from propagating into policy improvement steps. Similar to \cite{liu2021physics} and \cite{zeng2022data}, this ablation neither uses ensembling nor rollout scheduling to diversify or truncate rollouts in an adaptative manner. %In place of our training procedure that operates on consecutive prediction steps, we tried using single state transitions as in~\citep{zeng2022data} for the ablation but ultimately found it to be too unstable. 
    In contrast to our surrogate, the model is trained for a fixed number of gradient steps in each iteration. %Unless we monitor the approximation loss on a separate validation set, we expect our model to under- or overfit to patterns in the training data, dependent on the fixed number of gradient updates. Furthermore, we expect forecasts to propagate errors to model-free updates for the agent, destabilizing its improvement.
    \item \emph{Surrogate ablation:} In place of our residual surrogate, we use the full-state prediction model to approximate the temporal state evolution for the algorithm as presented in Section~\ref{sec:learning-surrogates}. %suggested that learning temporal residuals improves forecasts in the small data regime, whereas the {\fontfamily{qcr}\selectfont ICs Ablation} outperforms our model on broader time scales, although its reward estimates turned out worse overall. We expect that the algorithm will take longer to converge, since it updates the agent with forecasts suffering from larger errors in the small data regime due to it predicting states instead of temporal state changes.
\end{itemize}

\begin{figure}[t]
    \centering
    \includegraphics[width=1.0\columnwidth]{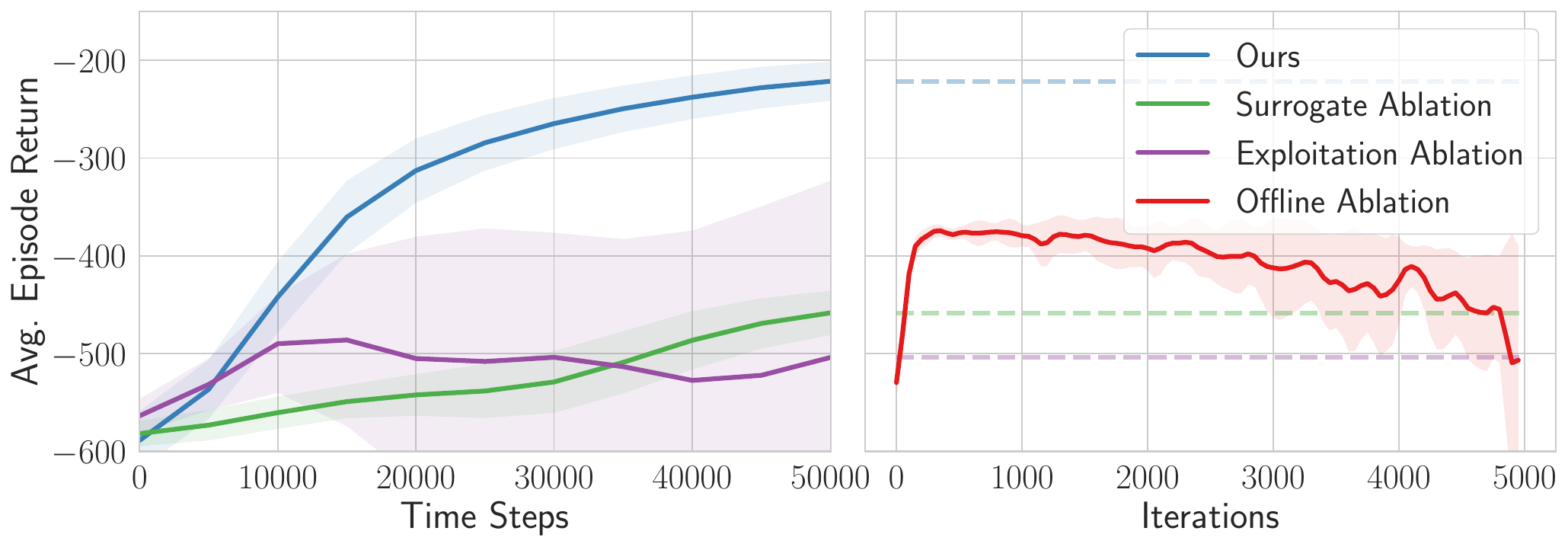}
    \caption{Average evaluation episode returns for our model-based approach compared to different ablations. Solid lines show the mean of three trials, while shaded regions denote the standard deviation among trials. All lines are smoothed using a Gaussian filter with $\sigma = 1$. The dotted lines on the right-hand side show the average performance of the left-hand algorithms after termination at $50,000$ time steps. Since our \emph{offline ablation} does not collect additional samples, we show its performance throughout iterations of the optimization.}
    \label{fig:learning-control:mbrl-ablations}
\end{figure}

Figure~\ref{fig:learning-control:mbrl-ablations} compares our implementation of the model-based methodology to its ablations. It shows that an improper treatment of compounding model errors often leads to unstable behavior optimization (purple line). Out of all methods we have tried for learning control strategies in our experiments, we found the \emph{exploitation ablation} to be the least stable, as characterized by its wide standard deviation. Implementing model ensembling and using a schedule to determine the length of model-based rollouts is seemingly essential to prevent the agent from taking advantage of model errors in the small data regime. The figure also shows that a similar issue pertains to our \emph{surrogate ablation} (green line). Although convergence is more stable overall, its approximate MDP cannot faithfully match the temporal evolution of the PDE in the small data regime so that behavior improvements of the agent for the surrogate do not generalize to the system. Specifically, as a result of the \emph{surrogate ablation} learning solution variables in place of temporal residuals, its forecasts sustain substantial errors at the beginning of data collection. 
%At the same time, it outperformed our surrogate in the offline experimental evaluation for time horizons of no less than 3.25 time units (13 discrete-time steps), that is, only for time spans longer than the extent of the rollouts our algorithm uses surrogates for.\todo{WE FOUND THAT LONGER MODEL-ROLLOUTS DO NOT WORK !!! -> SEE NEW RESULTS !!!} 
Our results also suggest that random exploration does not cover all relevant regions of the state space that policies visit during their optimization process (red line). After an initial phase of improvement, the behavior of our \emph{offline ablation} diverges more and more from the state distribution that the model used for training, increasing prediction errors and destabilizing the optimization once the policy begins exploiting them. Indeed, while trials are relatively stable at first, their standard deviation dilates with an increasing number of algorithm iterations.

\section{Discussion and outlook}
\label{sec:discussion}
We have seen that carefully constructed surrogate models are capable of increasing the sample efficiency by roughly one order of magnitude. 
Furthermore, we found that an overall more sophisticated approach to model-based reinforcement learning (online model adaptation, ensembling, curriculum learning, etc.) is beneficial to stabilize the convergence, and we hope that our findings showcase the appeal of adopting best practices.
However, due to the increased complexity, the performance of model-based RL is still inferior in situations where the model is comparatively easy to simulate.
For future work, it will thus be highly interesting to see whether simpler yet more efficient surrogate models can be utilized. At this point, suitable candidates appear to be GRUs instead of LSTMs (see, e.g., \cite{PTMS22}), sparse regreesion techinques to identify governing equations \cite{brunton2016discovering,KKB18}, and in particular the highly popular Koopman operator \cite{proctor2016dynamic,korda2018linear,peitz2019koopman}, as it allows us to learn linear models of nonlinear systems, which is very efficient both in terms of the required training data and the run time. Finally, it might be worth looking into recent prediction error results for these methods \cite{NPP+23,PB23,ZZ23} and see whether they can be transferred into guarantees for the RL process.
In addition, one can try to exploit system knowledge (in particular symmetries / invariances) in order to get smaller agents and thus to reduce the number of parameters that have to be trained \cite{BRV+19,PSC+23,GRS+23}.

\section*{Code availability}
The source code of the conducted experiments can be obtained freely under \textit{https://github.com/stwerner97/pdecontrol}.

\section*{Acknowledgment}

S.P.\ acknowledges financial support by the project ``SAIL: SustAInable Life-cycle of Intelligent Socio-Technical Systems'' (Grant ID NW21-059A), which is funded by the program ``Netzwerke 2021'' of the Ministry of Culture and Science of the State of Northrhine Westphalia, Germany. 
%, funded by the federal government of Northrhine-Westfalia, Germany.

% ####################################################################### %
\bibliographystyle{IEEEtran}
\bibliography{literature}

% ####################################################################### %

\appendix

\subsection{Neural network details}
\label{app:NN}
\begin{figure}
    \centering
    \includegraphics[width=1\columnwidth]{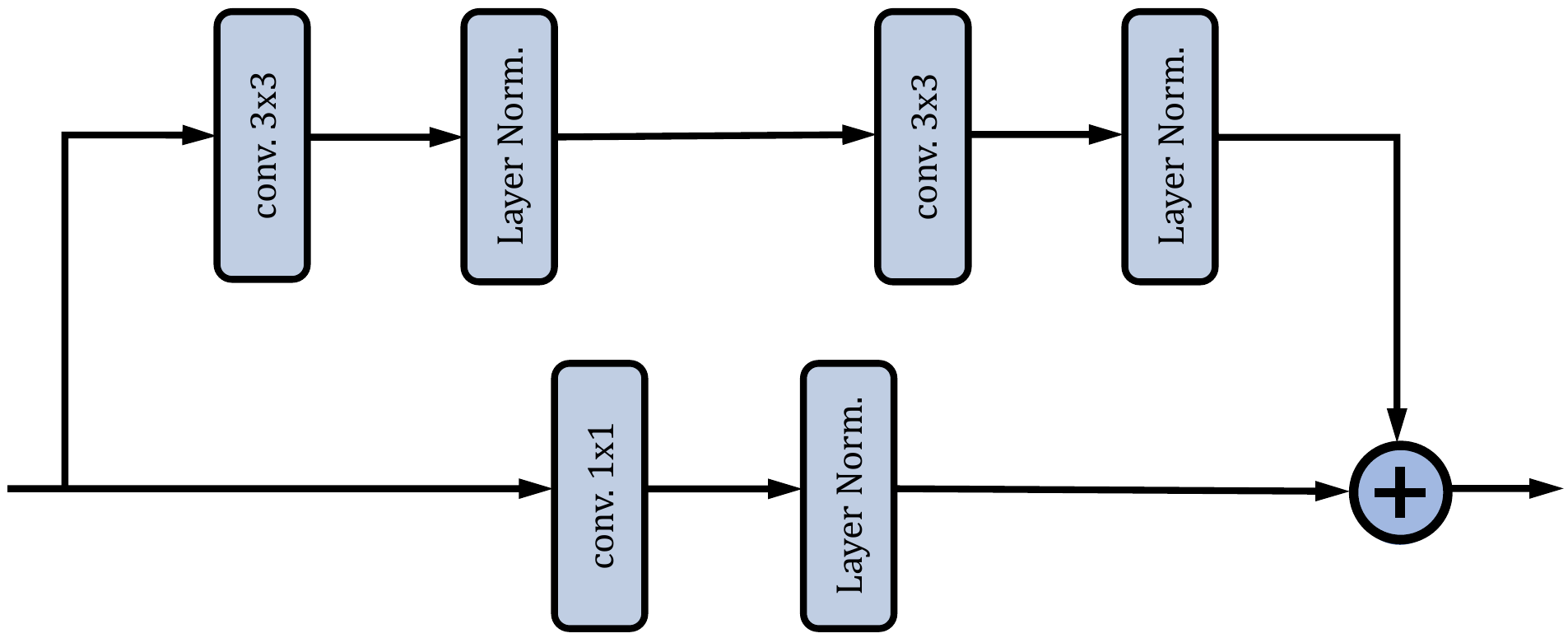}
    \caption{Residual block consisting of a skip connection (lower branch) downsampling the inputs with $1 \times 1$ convolutions and a main branch using $3\times 3$ convolutions. Layer normalization \cite{ba2016layer} is used to normalize the output activations for single input instances, which accelerates the training. The illustration is adopted from \cite{vahdat2020nvae}.}
    \label{fig:learning-surrogates:residual-cells}
\end{figure}

Our encoder network consists of residual cells similar to those proposed in \cite{vahdat2020nvae} for image generation, as they mitigate vanishing and exploding gradients at deeper layers of the architecture with so-called skip connections. The cells used in our work stack two $3\times3$ convolutional layers with \emph{SiLU} activation functions \cite{elfwing2018sigmoid} along the residual branch, see Figure \ref{fig:learning-surrogates:residual-cells} for an illustration. Similar to \cite{he2016deep}, we use a $1\times1$ strided convolution unit along the shortcut connection to downsample the input. Strides are also applied in the first $3\times3$ convolution of the residual branch to reduce the spatial dimensionality and dictate the degree to which the spatial extent is reduced. We typically half the spatial dimensionality with each stacked residual cell in the encoder network and use convolution instead of pooling layers for downsampling. 
%We further omit the \textit{Squeeze and Excitation}~\cite{hu2018squeeze} gating mechanism used in~\citep{vahdat2020nvae}, since its absence did not deteriorate accuracy in our experiments. 
Lastly, we use \emph{weight normalization} \cite{salimans2016weight} and \emph{layer normalization} \cite{ba2016layer} over \emph{batch normalization} \cite{ioffe2015batch} layers due to the well-known shortcomings of the latter during inference in recurrent structures.
For our decoder network $f_{\boldsymbol{\theta}_{\mathrm{dec}}}$, we use \textit{deconvolutions} \cite{zeiler2010deconvolutional,noh2015learning} without unpooling layers and with SiLU activation functions.
%We do not use unpooling layers in the decoder (similar to, e.g.,~\citep{clevert2015fast}). 
After the spatial extent of the physical domain has been restored, we introduce further convolutional layers to the architecture, which enhanced the performance in our experiments. %The kernel size of the final convolutional layer is set according to the maximum number of finite difference coefficients used in the numerical simulation of the temporal dynamics.

For the control term, we experimented with sharing the encoder network $f_{\boldsymbol{\theta}_{\mathrm{enc}}}$ between encoding states $\boldsymbol{u}_{\tau\Delta\tau}$ and forcing terms $\bphi_{\tau\Delta\tau}$.
However, since we only consider an additive control term in our application, we ultimately found training a separate encoder network $\boldsymbol{v}_{\tau} = f_{\psi_{\mathrm{enc}}}(\bphi_{\tau\Delta\tau})$ to be more effective. 
In general, the action encoder $f_{\psi_{\mathrm{enc}}}$ matches the model architecture of $f_{\boldsymbol{\theta}_{\mathrm{enc}}}$ but implements fewer convolutional filters due to the simple Gaussian shape of the exerted force in the case of the Kuramoto-Sivashinsky control problem considered here.

Besides parameter sharing, another benefit of the convolution operation is its ability to obey the boundary conditions. In the case of periodic boundaries, as studied in our work, a circular padding mechanism is applied at each border of the spatial domain. Similar to other inductive biases like spatial locality, encoding such prior system knowledge often improves data usage and generalization of forecasts. Using circular padding for all convolution operators in the encoder, decoder, and transition model helped significantly when learning the state evolution in our experiments.

\subsection{Implementation details on ensembles}
\label{app:ensembles}
Other works \cite{kurutach2018model,wang2019exploring,janner2019trust} alternate the dynamics model $f_{\boldsymbol{\theta}_{\ell}}$ used to predict state transition $\boldsymbol{\hat{s}}_{\tau + 1} = f_{\theta_{\ell}}(\boldsymbol{\hat{s}}_{\tau}, \boldsymbol{a}_{\tau} ; \, \boldsymbol{c}_{\tau, \ell})$ at each step. Owing to the fact that the memory unit of our model maintains a state of its past inputs, switching transition models after each step is only feasible if we invoke all models of the ensemble for each step. Otherwise, the memory kept as cell state $\boldsymbol{c}_{\tau, \ell}$ would be outdated. Saving computation, our work explores an alternative design choice and selects single models of ensemble $\{f_{\boldsymbol{\theta}_1}, \dots, f_{\boldsymbol{\theta}_{L_{\mathrm{ens}}}}\}$ and uses them throughout the entire model-based rollout, similar to what is considered in \cite{chua2018deep} for model predictive control. Our hope is that the use of an intermediate replay buffer~$\mathcal{D}_{\mathrm{model}}$ and training our model-free agent off-policy will have a similar decorrelation effect, since agents still update their parameters with artificial experience generated through several models at once. Indeed, our experiments show that there is no significant difference between both implementations for the decision-making problem studied in our work. The compute savings for model-based rollouts amount to a factor equal to the number of elites in the ensemble.

\subsection{Model-based rollouts}
\label{app:rollouts}

\begin{figure}
    \centering
    \includegraphics[width=.8\columnwidth]{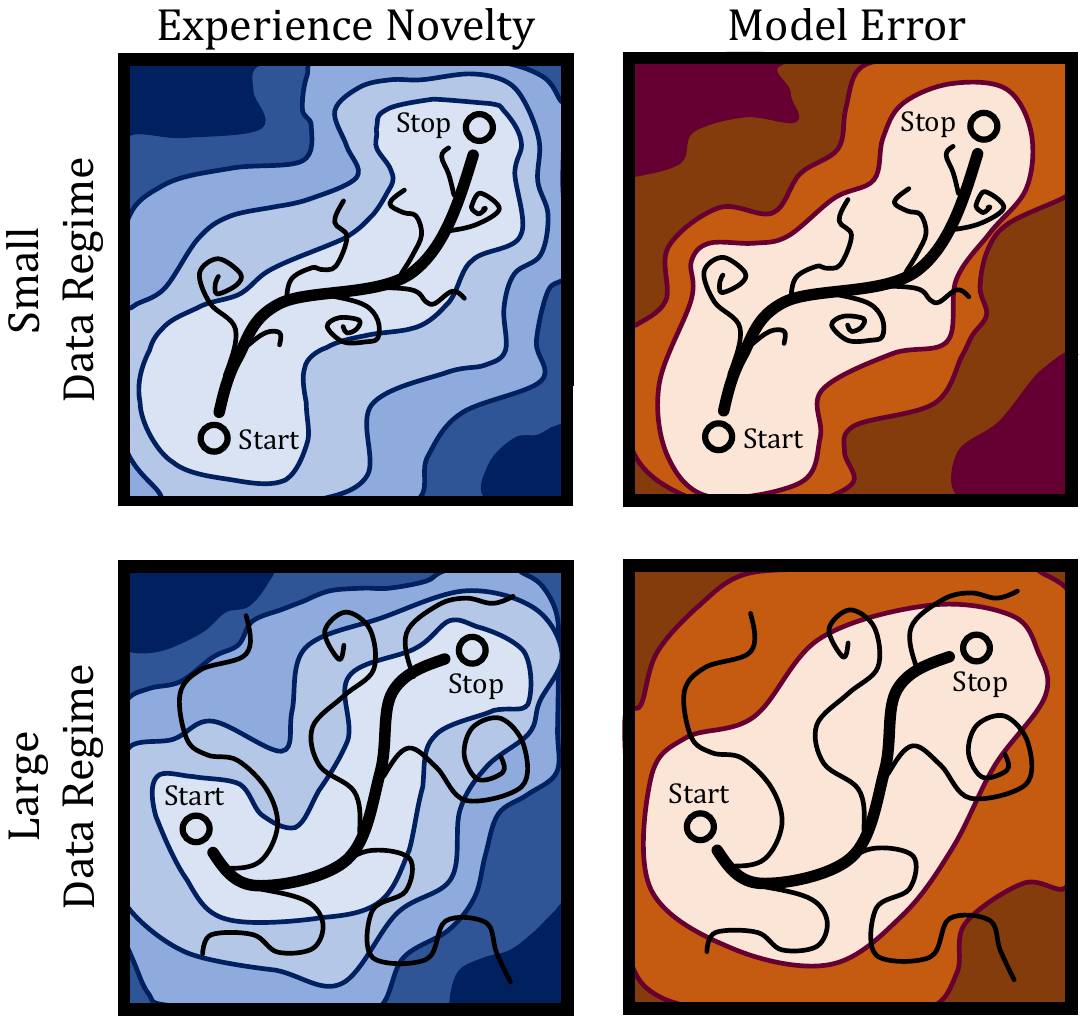}
    \caption{Intuition of rollout length scheduling. In the small data regime (top), the model does not generalize to unfamiliar regions of the state space (top right) so that rollouts must be truncated after few steps. In the large data regime (bottom), forecast can branch off collected data for several steps before compounding model errors become substantial (bottom right). In an ideal scenario, the model can generate unfamiliar experience for the agent (left-hand side), since it generalizes more efficiently.}
    \label{fig:learning-to-control:rollout-trade-offs}
\end{figure}

In order for model-based algorithms to succeed, planning must be advantageous to updating the behavior with existing samples or collecting additional data. On this matter, a recent study \cite{holland2018effect} formulates an essential condition for models to benefit the optimization. Specifically, unlike existing experience, a model has to generalize and has to be queried at unseen regions of the state space. A working assumption is thus that the model has to generalize quicker towards the global dynamics than the policy $\pi_{\boldsymbol{\omega}}$ can learn and generalize behavior. A prolonged rollout of $\pi_{\boldsymbol{\omega}}$ in the surrogate is thus inevitable for planning to generate samples unfamiliar to the agent \cite{holland2018effect}. Vice versa, with an increasing rollout length, model errors compound and could cause catastrophic learning updates to the agent \cite{van2019use}. As proposed in \cite{janner2019trust}, we adopt a scheduler for the rollout length $K_{\mathrm{rll}}$ in order to trade off compounding model errors against the novelty of samples. A rollout length scheduler deters the agent from exploiting model errors, specifically in the small data regime, where overfitting is often an issue of high-capacity function approximators. In general, we use our previous results for the valid prediction time of Section \ref{sec:learning-to-control-pdes} for guidance to limit the number of data collection steps. Our implementation uses a linear scheduler that increments the length $K_{\mathrm{rll}}$ up to a maximum number of steps. In Figure~\ref{fig:learning-to-control:rollout-trade-offs}, we illustrate the competing effects of sample novelty and compounding model errors in detail. The figure shows an ideal situation where the model quickly generalizes to other regions of the state space once more and more data becomes available (contour lines widening on the right-hand side), while the ability of the agent to generalize its behavior lags behind (marginal widening of contour lines on the left-hand side). Despite model-based rollouts increasing in depth, model-based rollouts continue to be accurate due to the generalizing model (bottom right) and collect data on state space regions unfamiliar to the agent (bottom left).

Overall, model-based rollouts proceed in the following way. After a sampled sequence of existing experience $(\boldsymbol{s}_{\tau - K_{\mathrm{tf}}}, \boldsymbol{a}_{\tau - K_{\mathrm{tf}}}, \dots, \boldsymbol{s}_{\tau}, \boldsymbol{a}_{\tau})\sim \mathcal{D}_{\mathrm{env}}$ is processed in teacher-forcing mode to obtain $\hat{\boldsymbol{s}}_{\tau}$, the policy~$\pi_{\boldsymbol{\omega}}$ selects an action~$\boldsymbol{a}_{\tau}$ as its control output. A dynamics model $f_{\boldsymbol{\theta}_{\ell}}$ of the ensemble then predicts the state transition $\hat{\boldsymbol{s}}_{\tau + 1} = f_{\boldsymbol{\theta}_{\ell}}(\hat{\boldsymbol{s}}_{\tau}, \boldsymbol{a}_{\tau} ; \, \boldsymbol{c}_{\ell, \tau})$. In practice, dynamics model~$f_{\boldsymbol{\theta}_{\ell}}$ compresses the state-action pair to a latent space using encoder networks $f_{\boldsymbol{\theta}_{\ell, \mathrm{enc}}}$ and $f_{\boldsymbol{\psi}_{\ell, \mathrm{enc}}}$, performs a time step using the convolutional LSTM $f_{\boldsymbol{\theta}_{\ell, \mathrm{fwd}}}$ and restores a state space representation with decoder network $f_{\boldsymbol{\theta}_{\ell, \mathrm{dec}}}$. Using the known reward function $\mathcal{R}$, we estimate the reward of the action as $\hat{r}_{\tau} = \mathcal{R}\left(\hat{\boldsymbol{u}}_{\tau\Delta\tau}, \bphi_{\tau\Delta\tau} \right)$ and insert the transition as an additional sample into dataset $\mathcal{D}_{\mathrm{model}}$. Rollouts are stopped manually after $K_{\mathrm{rll}}$ steps or conclude once a terminal state of the system is reached. The studied decision-making problem in our work does not define a terminal condition except for the fixed time limit of $T_{\mathrm{max}}$.

\subsection{Model training details}
\label{app:training}
Given sequences of system states~$\{\boldsymbol{s}_{\tau}\}_{\tau=0}^{K_{\mathrm{seq}}}$ and control outputs~$\{\boldsymbol{a}_{\tau}\}_{\tau=0}^{K_{\mathrm{seq}}-1}$ for training, we first use our surrogate model $f_{\boldsymbol{\theta}}$ to predict $\{\boldsymbol{\hat{s}}_{\tau}\}_{\tau=1}^{K_{\mathrm{seq}}}$. Afterwards, we train the model end-to-end while minimizing the mean squared error $\mathcal{L}_{\mathrm{MSE}} \propto \sum_i\sum_{\tau=1}^{K_{\mathrm{seq}}}\norm{\boldsymbol{\hat{s}}_{\tau}^{(i)} - \boldsymbol{s}_{\tau}^{(i)}}^2_2$ averaged over batches and sequence lengths. To address the slow convergence of the optimization and instabilities during training, we adopt teacher forcing \cite{williams1989learning} for guidance. In doing so, we split state sequence $\{\boldsymbol{s}_{\tau}\}_{\tau=0}^{K_{\mathrm{seq}}}$ into separate portions $\{\boldsymbol{s}_{\tau}\}_{\tau=0}^{K_{\mathrm{tf}}}$ and $\{\boldsymbol{s}_{\tau}\}_{\tau=K_{\mathrm{tf}}+1}^{K_{\mathrm{seq}}}$ and replace model predictions $\boldsymbol{\hat{s}}_{\tau + l}$ with states $\boldsymbol{s}_{\tau + l}$ in the first $K_{\mathrm{tf}}$ steps of the recurrent feedback loop to mitigate compounding errors propagating to later prediction steps at the beginning of our training process. After $K_{\mathrm{tf}}$ steps with teacher feedback, the model predicts sequence $\{\boldsymbol{\hat{s}}_{\tau}\}_{\tau=K_{\mathrm{tf}}+1}^{K_{\mathrm{seq}}}$ using its prior output $\boldsymbol{\hat{s}}_{\tau + l}$ as inputs for the succeeding prediction step $\boldsymbol{\hat{s}}_{\tau + l + 1}$.
In an attempt to reconcile the benefits of teacher forcing in terms of convergence speed with the necessity of consecutive prediction steps, we use a curriculum for training \cite{bengio2009curriculum}. At first, we keep the number of consecutive state transitions~${K_{\mathrm{seq}}} - K_{\mathrm{tf}}$ small to learn the encoding and decoding of state variables quickly. After that, we shift the focus towards learning temporal transitions and increase the sequence length ${K_{\mathrm{seq}}}$ towards the target sequence length $K_{\mathrm{max}} + K_{\mathrm{tf}}$, that is, predict $K_{\mathrm{max}}$ steps in feedback mode. 

Given a training set of~$N_{\mathrm{eps}}$ episodes composed of~$T_{\mathrm{max}} / \Delta\tau$ state transitions each, we subsample sequences $\{\boldsymbol{s}_{\tau}, \boldsymbol{a}_{\tau}, \dots, \boldsymbol{s}_{\tau + K_{\mathrm{seq}}}, \boldsymbol{a}_{\tau + K_{\mathrm{seq}}}\}$ starting at arbitrary points within episodes. In each training and validation epoch, we subsample $K_{\mathrm{subseq}}$ sequences equal to the number of consecutive segments in the dataset which are non-overlapping and of length $K_{\mathrm{seq}}$. While we subsample our training sequences anew after each epoch, the validation set remains fixed throughout the optimization procedure. 

We use the Adam optimizer \cite{kingma2014adam} with an adaptive learning rate (initial learning rate of $0.001$, momentum of $0.9$) to minimize the training loss $\mathcal{L}_{\mathrm{MSE}}$. Standard techniques, including gradient clipping (threshold of $ 0.5$) and regularization, are used. Using \textit{truncated backpropagation through time} \cite{jaeger2002tutorial}, we limit the maximum number of steps $K_{\mathrm{TBTT}}$ the gradient signal is backpropagated. 
% As indicated beforehand, we use gradient clipping at a threshold of $ 0.5$ and the Adam optimizer with an initial learning rate of $0.001$ and momentum of $0.9$ for training. 
All inputs as well as prediction targets are scaled to the unit interval or normalized based on statistics derived from the training set. In our experiments, learning normalized target values of the temporal residuals $\Delta\boldsymbol{s}_{\tau} / \Delta \tau = ( \boldsymbol{s}_{\tau + 1} - \boldsymbol{s}_{\tau}) / \Delta\tau$ at each step worked significantly better than backpropagating through reconstructed state variables $\boldsymbol{\hat{s}}_{\tau + 1} = \boldsymbol{\hat{s}}_{\tau} + \Delta\boldsymbol{\hat{s}}_{\tau}$. 

Table~\ref{tab:learning-surrogates:dynamics-model-architecture} shows the network architecture for our dynamics model. In essence, the state and action encoder networks compress the number of spatial dimensions of the state variable four-fold.

\begin{table}
    \centering
    \caption{Network architecture of our dynamics model.}
    \resizebox{1\columnwidth}{!}{
    \begin{tabular}{l|l|l|l|l|l|l}
      \toprule
    Network    & Layer & Filters &	Filter	&	Stride & 	Parameters	&	Output	\\
        &  &  &	 Size	&	 & 		&\\
        	\midrule
    State $\boldsymbol{s}$  & ---	&	---	&	--- & 	---	& ---	&	$1 \times 64$	\\
        	\midrule
   
    \multirow{3}{*}{State Encoder} &  Res.\ Block  	& 4 &	3	&	1	&	64  &	$4 \times 64 $ \\
   
    & Res.\ Block &	8    &  3	 &	2 &	320		& $8 \times 32 $ \\
   
    & Res.\ Block &	16	 &  3    &	2 &	1280	& $16 \times 16 $ \\
   
    \midrule
    
        Forcing Term $\bphi$  & ---	&	---	&	--- & 	---	& ---	& $1 \times 64$	\\
        	\midrule
   
    \multirow{3}{*}{Action Encoder} &  Res.\ Block  	& 2 &	3	&	1	&	20  &	 $2 \times 64 $ \\
   
    & Res.\ Block &	4    &  3	 &	2 &	80		&  $4 \times 32 $ \\
   
    & Res.\ Block &	6	 &  3    &	2 &	204	&  $6 \times 16 $ \\
    
          \midrule

    \multirow{1}{*}{Transition Model} &  Conv. LSTM	& 16 &	3	&	1	&	4800 &	$16 \times 16 $ \\

          \midrule
    
    \multirow{4}{*}{State Decoder} &  DeConv.  	& 16  & 	3	&	2	&	784 &	$16 \times 32 $ \\
   
    & DeConv. &	8    &  3	 &	2 &	392		& $8 \times 64 $ \\
   
    & Res.\ Block &	4	 &  5    &	1 &	176	& $4 \times 64$   \\
    
    & Conv. &	1	 &  7    &	1 &	13	& $1 \times 64$   \\
   
    \bottomrule
    \end{tabular}
    }
    \vspace{5pt}
    \label{tab:learning-surrogates:dynamics-model-architecture}
\end{table}

\subsection{Study on the neural network sizes}
\label{app:NNsize}
We here study several variants of our dynamics model with an increasing number of parameters. In doing so, we evaluate surrogates similar to the encoder, decoder and transition model outlined in Table~\ref{tab:learning-surrogates:dynamics-model-architecture}, but scale the width $\beta$ of each network, which defines the exponential number of channels $\beta^N$ of its layers in terms of their depth $N$ \cite{tan2019efficientnet}. 
%In case of a state encoder with three layers, for example, a network of width $\beta = 4$ implements residual blocks of $4, 16$ and $64$ channels. Vice versa, the state decoder outputs $64$, $16$ and $4$ channels before the final convolution block. 
%The number of filters for the convolutional \gls{lstm} is in line with the number of channels of the encoded state, while the action encoder remains unchanged. 
We evaluate the predictive capabilties of surrogates with network widths and overall number of parameters outlined in Table~\ref{tab:learning-surrogates:width-model-parameters}. 
In this setting, the network in our previous study (which we are also going to use in the next section) corresponds to $\beta \approx 3$.

\begin{table}[h]
    \centering
    \caption{Number of learnable parameters at increasing model widths.}
    \begin{tabular}{c|cccccc}
      \toprule
    Width $\beta$    & 1.5 & 2.0 &	2.5	&	3.0 & 	3.5	&	4.0 \\
        	\midrule
    Model Parameters  & 2612	&	4137	&	7636 & 	17542	& 36409	& 77443 \\
    \bottomrule
    \end{tabular}
    \vspace{5pt}
    \label{tab:learning-surrogates:width-model-parameters}
\end{table}

\begin{figure}[h]
    \centering
    \includegraphics[width=1.0\columnwidth]{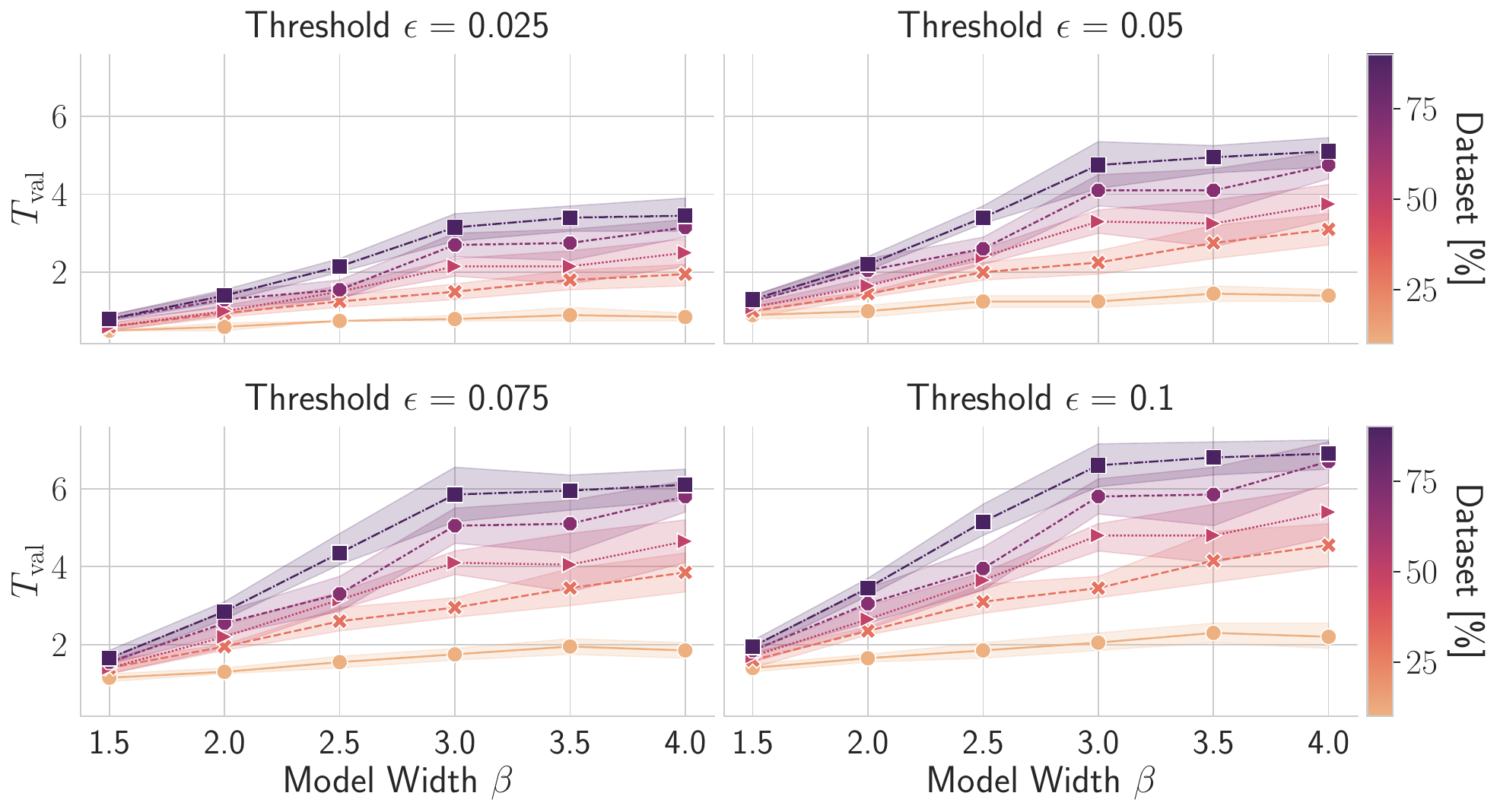}
    \caption{Valid prediction time of models at increasing model widths and several thresholds. Solid lines depict the mean of five folds, while the shaded regions show the 95\% confidence interval.}
    \label{fig:learning-surrogates:vpt-capacity}
\end{figure}

In order to determine the time span throughout which forecasts match the spatio-temporal evolution of state variables and attributed rewards, we compute the \textit{valid prediction time} as the farthest point in time $T_{\mathrm{vpt}}$ with normalized mean squared errors of state and reward being consistently below a fixed threshold $\epsilon$. 
Therefore, $T_{\mathrm{vpt}}$ gives a coarse estimate on the number of steps we expect to be safe in the context of model-based reinforcement learning. In Figure~\ref{fig:learning-surrogates:vpt-capacity}, we illustrate the valid prediction time of models at increasing network widths for a number of thresholds and dataset sizes given for training. Surprising to us, models of larger network widths are not susceptible to overfitting if only few snapshots are available, with our model at width $\beta = 4.0$ outperforming the network at width $\beta = 1.5$ consistently. (In contrast, we observed an inverse trend on fully connected architectures.) As a matter of fact, the valid prediction time consistently increases with growing network width but plateaus marginally after $\beta=3.0$. 

Since computation time increases at an exponential rate with the width of a neural network \cite{tan2019efficientnet}, we fix $\beta \approx 3.0$ for subsequent experiments. Intermediate experiments in our work suggested that increasing model capacity also improves data efficiency and overall performance of learned control strategies. Still, a comprehensive evaluation of our algorithm for model-based control turned out to be too resource-intensive and time-consuming for larger dynamics models.

\end{document}